\documentclass[journal]{IEEEtran}
\usepackage{cite}
\usepackage{graphicx}
\usepackage{bm}
\usepackage{epsfig}
\usepackage{amsmath}
\usepackage{array}
\usepackage{multirow}
\usepackage{color}
\usepackage{amsmath}
\ifCLASSINFOpdf
\else
\fi
\begin{document}
%
\title{{CDPM: Convolutional Deformable Part Models for Semantically Aligned Person Re-identification}}
%
%
%

\author{Kan Wang,
        Changxing Ding,
        Stephen J. Maybank,
        and Dacheng Tao,

%
\thanks{Kan Wang and Changxing Ding are with the School of Electronic and Information Engineering,
South China University of Technology,
381 Wushan Road, Tianhe District, Guangzhou 510000, P.R. China
(e-mail: eekan.wang@mail.scut.edu.cn; chxding@scut.edu.cn).}
\thanks{Stephen J. Maybank is with the Department of Computer Science and Information Systems, Birkbeck College, London WC1E 7HX, UK (e-mail: sjmaybank@dcs.bbk.ac.uk).}
\thanks{Dacheng Tao is with the UBTECH Sydney Artificial Intelligence Centre and the
School of Computer Science, in the Faculty of Engineering, at The University of Sydney, 6 Cleveland St, Darlington, NSW
2008, Australia (e-mail: dacheng.tao@sydney.edu.au).}}

\maketitle

\begin{abstract}
 Part-level representations are essential for robust person re-identification. However, common errors that arise during pedestrian detection frequently result in severe misalignment problems for body parts, which degrade the quality of part representations. Accordingly, to deal with this problem, we propose a novel model named Convolutional Deformable Part Models (CDPM). CDPM works by decoupling the complex part alignment procedure into two easier steps: first, a vertical alignment step detects each body part in the vertical direction, with the help of a multi-task learning model; second, a horizontal refinement step based on attention suppresses the background information around each detected body part. Since these two steps are performed orthogonally and sequentially, the difficulty of part alignment is significantly reduced. In the testing stage, CDPM is able to accurately align flexible body parts without any need for outside information. Extensive experimental results demonstrate the effectiveness of the proposed CDPM for part alignment. Most impressively, CDPM achieves state-of-the-art performance on three large-scale datasets: Market-1501, DukeMTMC-ReID, and CUHK03.
\end{abstract}


%
\IEEEpeerreviewmaketitle

\section{Introduction}
\IEEEPARstart{P}{erson} re-identification (ReID) refers to the recognition of one pedestrian's identity from images captured by different cameras. Given an image containing a target pedestrian (i.e., the query), a ReID system attempts to search a large set of pedestrian images (i.e., the gallery) for images that contain the same pedestrian. ReID has attracted substantial attention from both academia and the industry due to its wide-ranging potential applications, which include e.g. video surveillance and cross-camera tracking \cite{ristani2018features}. However, due to the large number of uncontrolled sources of variation, such as dramatic changes in pose and viewpoint, complex variations in illumination, and poor image quality, ReID remains a very challenging task.

\begin{figure}
  {\centerline{\includegraphics[width=0.5\textwidth]{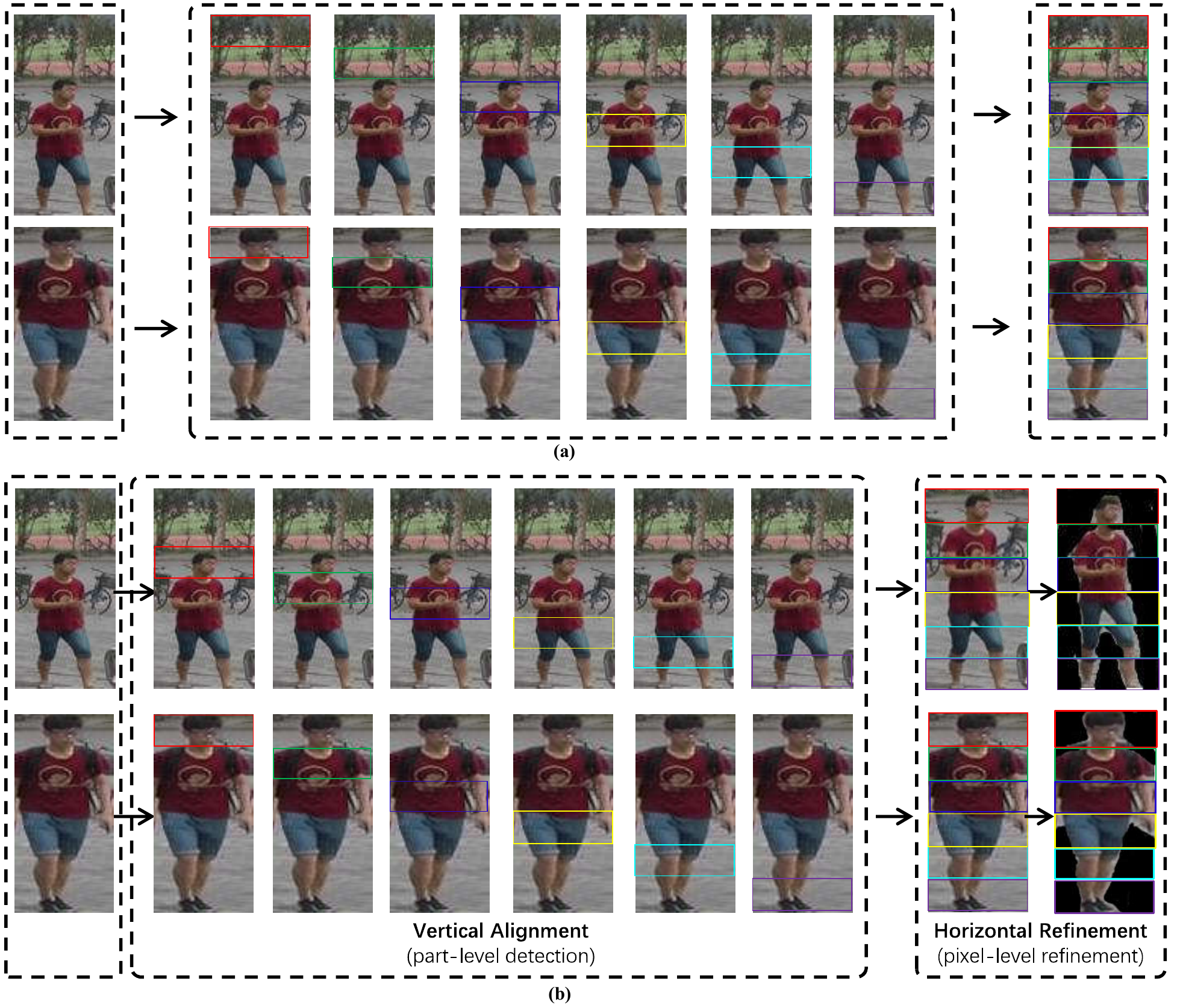}}
  \caption{(a) The pipeline of the baseline model based on uniform division \cite{sun2017beyond}. The severe part misalignment problem dramatically degrades the quality of the part-level representations. (b) The inference pipeline of CDPM. It decouples the part alignment problem into two easier steps, i.e., a vertical alignment step and a horizontal refinement step. In this way, body parts are semantically aligned across images. Best viewed in color.}
  \label{img1}
  }
\end{figure}

The key to a robust ReID system lies in the quality of pedestrian representations. Many approaches \cite{zheng2017person,zheng2016person} attempt to directly extract holistic-level features from the whole image. These approaches typically suffer from overfitting problems \cite{Yao2017Deep}. Recently, part-level representations have been proven to be highly discriminative and capable of achieving state-of-the-art performance \cite{wang2018learning,li2018harmonious,sun2017beyond,fu2018horizontal,Part-based2017tip}. {However, as illustrated in Fig. \ref{img1}(a), the location of each body part varies in images due to errors in pedestrian detection \cite{zhang2017alignedreid, zheng2018pedestrian}. Consequently, the extracted part-level representations are not semantically aligned across images, meaning that they are not directly comparable for ReID purposes.}

One intuitive strategy that could be adopted to resolve this issue involves directly detect body parts using additional tools, e.g., keypoints produced by pose estimation algorithms \cite{zhao2017spindle,su2017pose}, in both training and testing stages. However, the predictions made by these tools may not be sufficiently reliable, as they are usually trained on databases containing images that were captured under different conditions from images in ReID datasets. Another popular strategy involves detecting body parts via attention models that are seamlessly integrated into the ReID architecture \cite{li2018harmonious,li2017learning,lan2017deep,zhao2017deeply}. However, these attention models are optimized using the ReID task only; therefore, they are unable to provide explicit guidance for part alignment.

Accordingly, in this paper, we propose a novel framework for part alignment. By providing the training stage with {{a minimal extra annotation (the upper and lower boundaries of the pedestrian) that is automatically detected}}, we are able to factorize the complicated part alignment problem into two simpler and sequential steps, i.e., a vertical alignment step that detects body parts in the vertical direction, and a horizontal refinement step which suppresses the background information around each detected part, as illustrated in {{Fig. \ref{img1}(b)}}.

Based on the above idea, we introduce a novel end-to-end model named Convolutional Deformable Part Models (CDPM), which can both detect flexible body parts and extract high-quality part-level representations. CDPM is built on a popular convolutional neural network (CNN) as backbone and further constructs three new modules, i.e., a feature learning module that extracts part-level features, a vertical alignment module that detects body parts in the vertical direction via multi-task learning, and a horizontal refinement module which is based on the attention mechanism.

Different channels in a CNN describe different visual patterns \cite{Yao2017Deep, liu2017hydraplus, zhang2018occluded, ding2017trunk}, i.e., different body parts in the ReID context. In other words, channel-wise responses indicate hints of the location for each part. In light of the above, CDPM succinctly integrates these three modules, based on the output of the same backbone model. In the inference stage, the vertical alignment module and the horizontal refinement module run sequentially for part alignment, followed by high-quality feature extraction from the aligned parts.

The effectiveness of the proposed method is systematically evaluated on three popular ReID databases, i.e., Market-1501 \cite{Zheng2015Scalable}, DukeMTMC-reID \cite{zheng2017unlabeled}, and CUHK03 \cite{li2014deepreid}. Experimental results demonstrate that CDPM consistently achieves superior performance and outperforms other state-of-the-art approaches by a considerable margin.

We summarize the contributions of this work as follows:
\begin{itemize}
\item We formulate the novel idea of decoupling the body-part alignment problem into two orthogonal and sequential steps, i.e., a vertical detection step and a horizontal refinement step. These two steps establish a novel framework for the learning of high-quality part-level representations. To the best of our knowledge, this is the first attempt to solve the misalignment problem by means of decomposition into orthogonal directions.

\item Under the \emph{divide-and-conquer} formulation, we propose a succinct CDPM architecture that integrates representation learning and part alignment through sharing of the same backbone model. In particular, the vertical alignment module is realized by an elaborately designed multi-task learning structure.

\item Extensive evaluation on three large-scale datasets demonstrate the superiority of the proposed CDPM. We further conduct comprehensive ablation study to enable analysis of the effectiveness of each component of CDPM.
\end{itemize}

The remainder of this paper is organized as follows. We first review the related works in Section II. Then, we describe the proposed CDPM in more detail in Section III. Extensive experimental results on three benchmarks are reported and analyzed in Section IV, after which the conclusions of the present work are outlined in Section V.

\section{Related Work}
\subsection{Person Re-identification}
Prior to the prevalence of deep learning, approaches to ReID could be divided into two distinct categories, namely, feature engineering methods \cite{varior2016learning,zhao2013unsupervised} and metric learning methods \cite{chen2015relevance,Discriminant2017tip,Relevance2015tip,kernel2019tip,metric2018tip}.
Over the past few years, deep learning-based approaches \cite{newxiao2016learning,Yao2017Deep, newchen2018group,wang2018learning,li2018harmonious,sun2017beyond,shen2018end,fu2018horizontal,Part-based2017tip, zhao2017spindle,su2017pose,Learning2018tip,Video2019tip,Robust2017tip,Ranking2016tip,shen2018deep, End2017tip, newwang2018person, deng2018image} have come to dominate in the ReID community. Many works target the learning of discriminative representations from the holistic image directly. One common strategy involves training deep models to learn ID-discriminative embedding (IDE) as an image classification task \cite{zheng2017person}. Moreover, the quality of image representations can be enhanced using metric learning-based loss functions, such as contrastive loss \cite{Varior2016A}, triplet loss \cite{hermans2017defense}, and quadruplet loss \cite{chen2017beyond}. Other works train deep models using a combination of different types of loss functions \cite{macus2018,chen2017multi}.

The holistic image-based approaches typically suffer from overfitting problems \cite{Yao2017Deep}. To relieve this issue, part-based approaches \cite{Yao2017Deep,wang2018learning,li2018harmonious,sun2017beyond,fu2018horizontal,li2017learning} have been proposed in order to learn discriminative image representations. However, due to errors that commonly arise in pedestrian detection context, the location of each body part varies in different images. When considering strategies for handling this part misalignment problem, existing approaches can be grouped into three categories.

\subsubsection{Pre-defined Part Location-based Methods}
These approaches extract part-level features from patches \cite{ahmed2015improved} or horizontal stripes \cite{wang2018learning,fu2018horizontal,Cheng2016Person} of pre-defined locations. For example, Cheng \emph{et al.} \cite{Cheng2016Person} uniformly divide a pedestrian image into four horizontal stripes, from which part-level features are then extracted. Wang \emph{et al.} \cite{wang2018learning} also partition one image into horizontal stripes. It mitigates the part misalignment problem by extracting multi-granularity part-level features. However, as the above methods usually assume that the misalignment problem is moderate; they may therefore encounter difficulties when handling cases of severe misalignment.

\subsubsection{Outside Information-based Methods}
These methods align body parts through the use of outside information, e.g., masks produced by human parsing tools \cite{song2018mask,tian2018eliminating,kalayeh2018human} or key points detected by pose estimation algorithms \cite{zhao2017spindle,su2017pose,xu2018attention}. In these approaches, outside information is usually required in both the training and testing stages \cite{zhao2017spindle,su2017pose, kalayeh2018human}. The {downsides} to these methods are first, there is additional computational cost, second, the accuracy of part alignment depends on the performance of the outside tools, which are usually trained on databases made up of images captured under conditions that are significantly different from ReID databases.

\subsubsection{Attention Model-based Methods}
These methods learn to directly predict bounding boxes or soft masks for body parts from feature maps produced by ReID networks, without making use of any additional supervision \cite{li2018harmonious,li2017learning,zhao2017deeply,liu2017hydraplus}. For example, Li \emph{et al.} \cite{li2018harmonious} designed a hard regional attention model that is able to predict bounding boxes for each body part. In comparison, Zhao \emph{et al.} \cite{zhao2017deeply} proposed to predict a set of soft masks. Element-wise multiplication between one soft mask and each channel of feature maps is used to produce part-level features. However, the lack of explicit supervision of part alignment may cause difficulties during the optimization of these attention models.

The proposed CDPM approach improves the accuracy of part alignment by introducing a minimal amount of extra supervision in the training stage, through which the complicated part alignment problem can be decomposed into two separate and simpler steps. Therefore, compared with the attention-based methods, the optimization difficulty associated with part alignment is significantly reduced. Compared with the second category of methods, the utilized annotations are more robust. Besides, CDPM does not require any outside information in the testing stage; therefore it is easier to use in practice.

\subsection{Part-based Object Detection}
Prior to the emergence of deep learning as a widespread phenomenon, the Deformable Part Model (DPM) was one of the most popular methods for object detection. In both DPM \cite{felzenszwalb2008discriminatively} and its deep versions \cite{ouyang2015deepid, savalle2014deformable,girshick2015deformable}, part detection is performed as an auxiliary task to promote detection accuracy. Over the past few years, region proposal-based methods \cite{girshick2015fast,ren2017faster} have become more popular. Unlike DPM methods, region proposal-based methods usually detect the whole object directly rather than engaging in explicit part detection.

By contrast, the proposed method aims to detect flexible parts only, as the coarse location of the whole body is already known. From this perspective, our method bears more similarity to DPM than to region proposal-based methods. Since our proposed method is based on CNN, we name it Convolutional Deformable Part Models (CDPM).

\section {Convolutional Deformable Part Models}
\subsection{Problem Formulation}
As illustrated in Fig. \ref{img1}, we decouple the complex part alignment problem into two separate and sequential steps, i.e., a vertical alignment step that locates each body part in the vertical direction, and a horizontal refinement step which suppresses the background information around each body part.

The first step is more challenging. This is because, firstly, the whole image is searched for each body part; secondly, there is usually no clear boundary between adjacent parts. We meet the above challenges by providing a minimal extra annotation as a form of auxiliary supervision in the training stage.

\begin{figure}
  {\centerline{\includegraphics[width=0.5\textwidth]{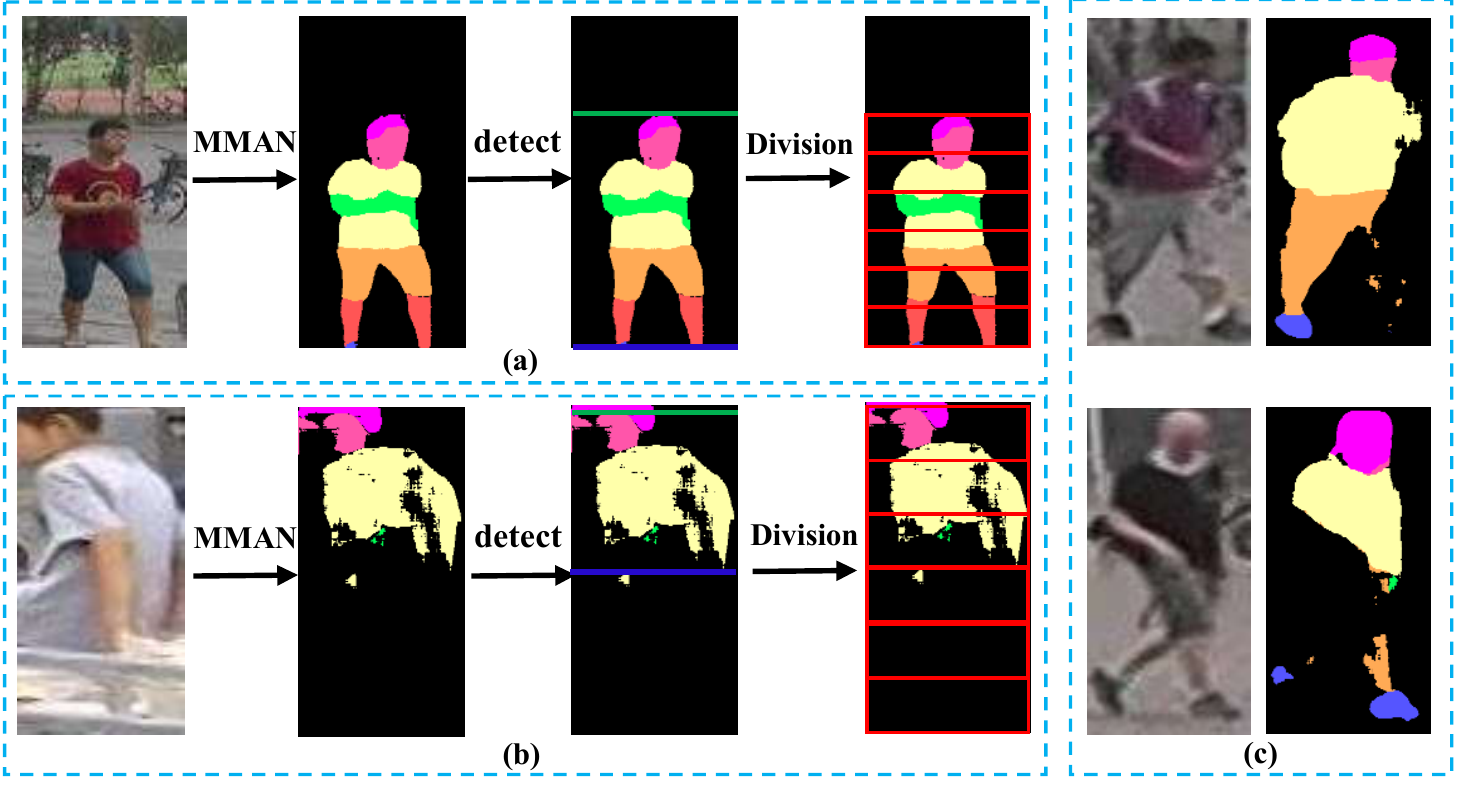}}
  \caption{ (a) The automatically detected upper and lower boundaries for a pedestrian, which are denoted as the green line and the blue line, respectively. The location of each part is inferred through the uniform partition between the two boundaries. (b) Severe part-missing problem is detected by counting the number of pixels for the head and the lower-part of the body, respectively. Best viewed in color. (c) The parsing results may not be sufficiently reliable for low-quality images.}
  \label{img2}
  }
\end{figure}

As shown in Fig. \ref{img2}(a), the upper and lower boundaries of pedestrians are automatically detected by Macro-Micro Adversarial Network (MMAN) \cite{luo2018macro}. The upper and lower boundaries of pedestrians are obtained according to the following rules. First, the seven classes obtained by MMAN\cite{luo2018macro} are merged into three categories, i.e., the head (including the hair and face), the upper body (consisting of the upper-clothes and arms), and the lower part of the body (comprises lower-clothes, legs and shoes). Second, the upper boundary of the pedestrian is set as the upper boundary of the head, while the lower boundary of this pedestrian is defined as the lower boundary of the lower part of the body. Besides, as shown in Fig. \ref{img2}(b), severe part-missing problem is detected by counting the number of pixels for the head and the lower-part of the body, respectively. If the size of either of them is smaller than a pre-set threshold (e.g., 1280 pixels in our implementation), we regard this image as having a severe part-missing problem. Since MMAN may fail to produce reliable results (Fig. \ref{img2}(c)), it is suboptimal to directly utilize the location of body parts returned by MMAN.

The equal partition between these two detected boundaries produces the area of each part in the vertical direction. It is worth noting here that the annotation is not required in testing; therefore, the detected annotation can be regarded as a kind of privileged information \cite{vapnik2015learning}. Moreover, the second step is comparatively much easier; therefore, it is not provided with any extra information.

\begin{figure*}
  \centerline{\includegraphics[width=1.0\textwidth]{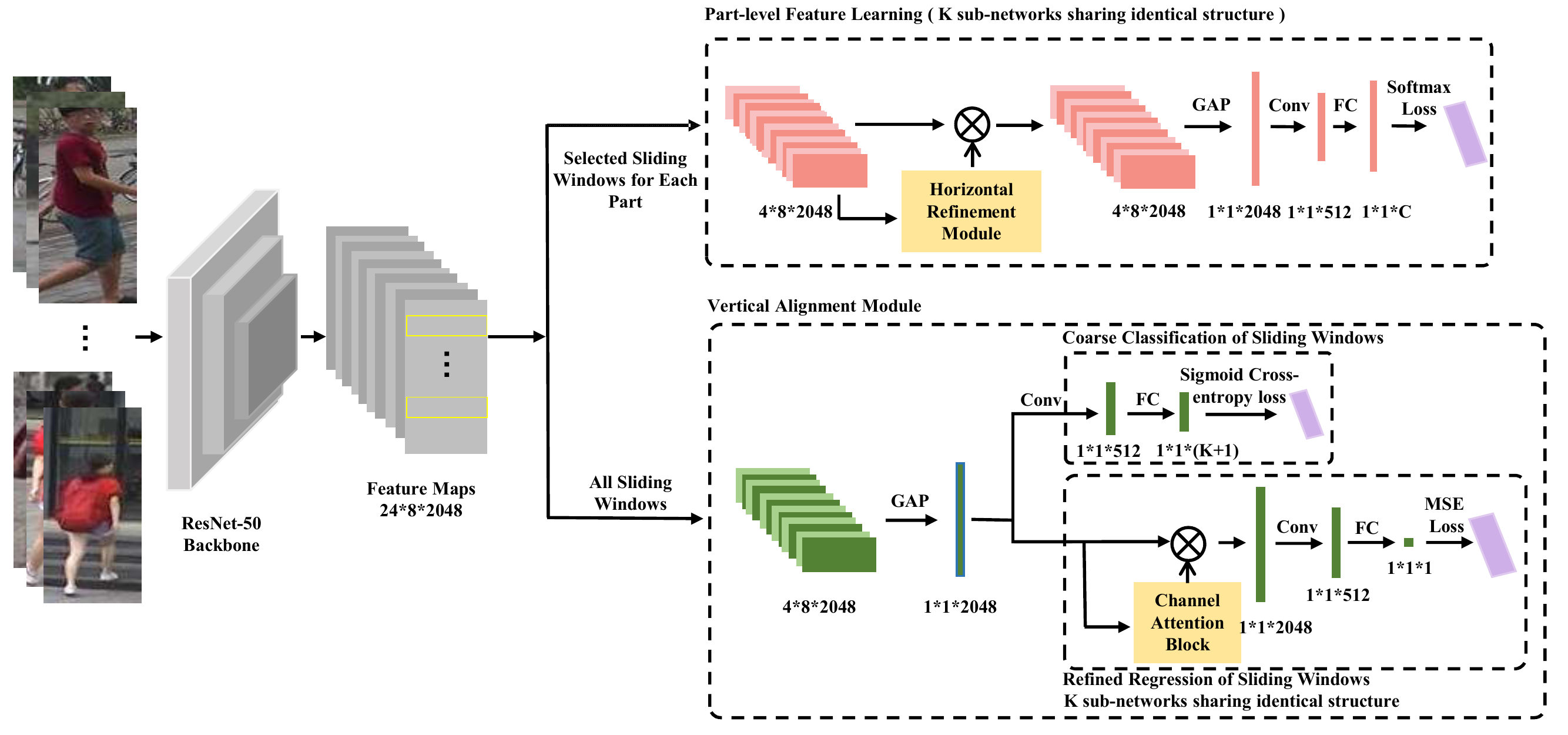}}
  \caption{ Architecture of the proposed CDPM. Based on the ResNet-50 backbone model, CDPM constructs three new modules, i.e., the feature learning module including \emph{K} part-level branches, the vertical alignment module, and the horizontal refinement module. In the inference stage, the vertical alignment module receives \emph{R} sliding windows for each image and selects one optimal sliding window for each part; the selected sliding window indicates the location of each part in the vertical direction. The horizontal refinement module further reduces the interference of background information in the selected sliding window. These two modules work together to achieve the goal of part alignment, thereby enabling the part-level feature learning branches to learn robust features.}
  \label{img3}
\end{figure*}

Based on the above ideas, we propose our novel CDPM model for joint part feature learning and part alignment. As illustrated in Fig. \ref{img3}, CDPM is built on the ResNet-50 backbone model \cite{he2016deep}. Similar to \cite{sun2017beyond}, we remove the last spatial down-sampling operation in ResNet-50 so as to increase the size of the output feature maps. Based on these output feature maps, we go on to construct three new modules, i.e., the feature learning module for part-level feature extraction, the vertical alignment module based on multi-task learning, and the horizontal refinement module based on attention. These three modules work collaboratively together to align body parts and further learn high-quality part-level representations.

\subsection{Feature Learning Module} The feature learning module is based on one recent work named Part-based Convolutional Baseline (PCB) \cite{sun2017beyond}. As illustrated in Fig. \ref{img3}, the feature learning module incorporates \emph{K} part-level feature learning branches, each of which learns part-specific features. These \emph{K} branches all have an identical structure, i.e., one Global Average Pooling (GAP) layer, one $1 \times 1$ convolutional layer, and one classification layer. In the training stage, the location of each part can be inferred via the provided annotations pertaining to the upper and lower pedestrian boundaries by means of uniform partition (Fig. \ref{img2}(a)). If the upper or lower boundary is not provided, e.g., in cases where they are invisible due to part-missing problems (Fig. \ref{img2}(b)), we uniformly divide the whole image in the vertical direction. In the testing stage, the location of each part is determined via the proposed part alignment method.

Each of the \emph{K} part-level features is optimized as a multi-class classification task using the softmax loss function. The loss function for the $k$-th part is formulated as follows:
\begin{equation}
\mathcal{L}_{p}^{k} = - \frac{1}{N}\sum_{i = 1}^{N}  \log \frac{e^{{{\bf{w}}_{y_{i}}^{k}}^{T}{\bf{z}}_{i}^{k}+{b}_{y_{i}}^{k}}}{\sum_{j =
1}^{C}{e^{{{\bf{w}}_{j}^{k}}^{T}{\bf{z}}_{i}^{k}+{b}_{j}^{k}}}},
\end{equation}
where ${\bf{w}}_{j}^{k}$ is the weight vector for class $j$, while ${b}_{j}^{k}$ is the corresponding bias term, \emph{C} denotes the number of classes in the training set, and $y_{i}$ and ${\bf{z}}_{i}^{k}$ represent the label and the $k$-th part-level feature for the $i$-th image in a batch, respectively. Therefore, the overall loss function for the feature learning module is as follows:
\begin{equation}
\mathcal{L}_{f} = \sum_{k = 1}^{K}\mathcal{L}_{p}^{k}.
\end{equation}

\begin{figure}
  {\centerline{\includegraphics[width=0.5\textwidth]{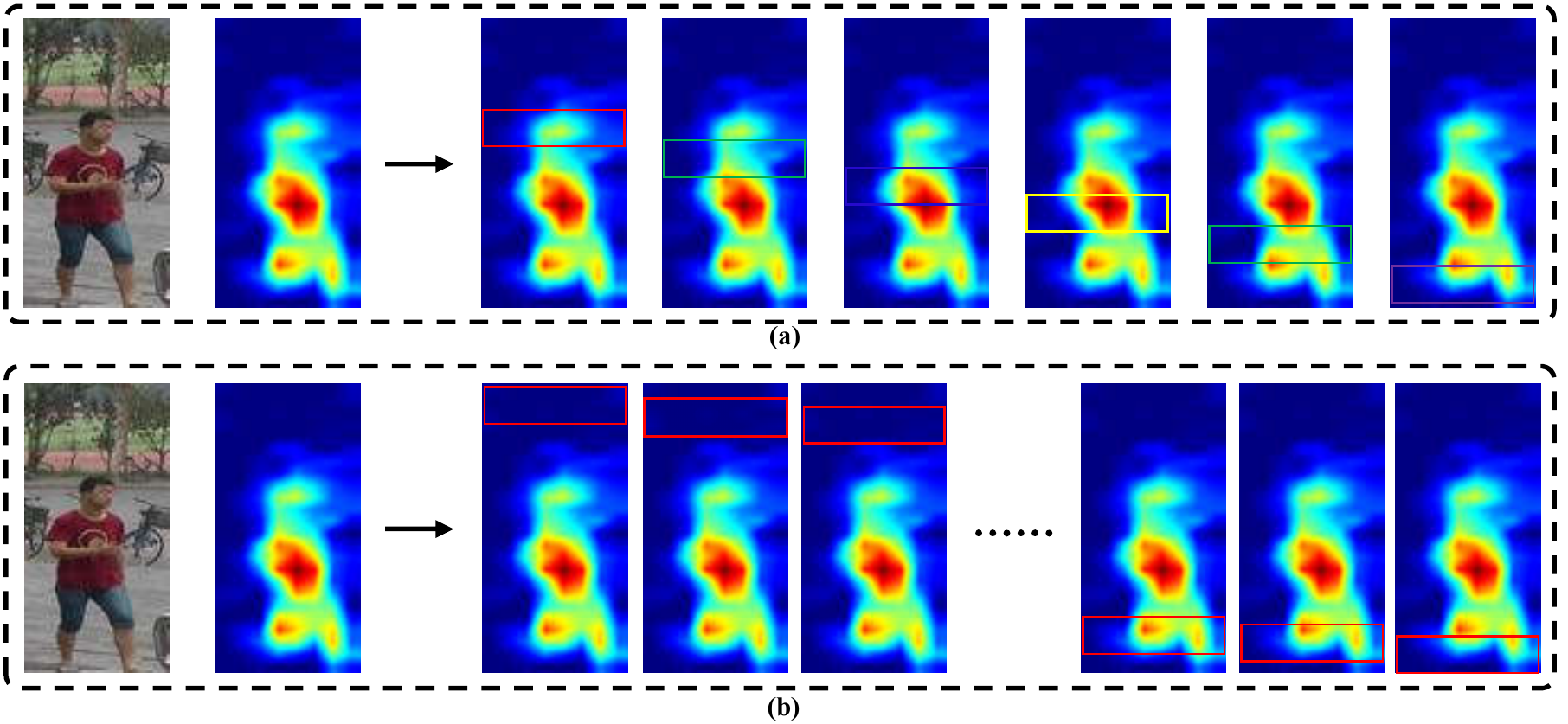}}
  \caption{(a) The \emph{K} selected sliding windows for the part-level feature learning modules. One sliding window is selected for each body part. (b) All sliding windows are utilized to train the vertical alignment module; there are 21 sliding windows for feature maps of size $24 \times 8$.}
  \label{img4}
  }
\end{figure}

\subsection{Vertical Alignment Module}
Different channels in the feature maps produced by the ReID network describe different body parts \cite{Yao2017Deep, zhang2018occluded}. This indicates that the channel-wise responses can provide hints as to part location. We therefore design a detection module to locate body parts in the vertical direction, based on the output of the backbone model only. In the interests of simplicity, we divide the output of the backbone model into $R$ sliding windows, each with fixed height and width. We then select one sliding window for each part using the proposed vertical alignment module. In this paper, the size of output of the backbone model is $24 \times 8 \times 2048$, where these three dimensions denote height, width, and channel number, respectively. The size of each sliding window is set as $4 \times 8 \times 2048$; therefore $R$ is equal to 21 for each image (Fig. \ref{img4}(b)).

Furthermore, inspired by Faster R-CNN \cite{ren2017faster}, we process the sliding windows by means of multi-task learning, i.e., coarse classification of all sliding windows, and the refined regression of sliding windows to their respective ground-truth locations. These two tasks share only one GAP layer. It is worth noting that we only utilize images whose upper and lower boundaries are both visible during the training for this module.

\subsubsection{Coarse Classification of Sliding Windows}
This task classifies all sliding windows to their corresponding parts or the background category. To do so, it incorporates one $1 \times 1$ convolutional layer and one fully connected (FC) layer. The important parameters for the two layers are outlined in Fig. \ref{img3}. The output of the FC layer is (\emph{K}+1)-dimensional, which denotes \emph{K} parts and the background category. As each of these sliding windows may overlap with two adjacent parts, their ground-truth labels are soft rather than one-hot. We therefore optimize the classification task using the sigmoid cross-entropy loss, which can be formulated as $\mathcal{L}_{c} = $
\begin{equation}
-\frac{1}{NR} \sum_{i = 1}^{N}\sum_{r = 1}^{R} \sum_{k = 1}^{K+1} [ y_{i}^{r(k)} \log \hat{y}_{i}^{r(k)} +
(1- y_{i}^{r(k)}) \log(1- \hat{y}_{i}^{r(k)}) ],
\end{equation}
where $y_{i}^{r(k)}$ is the ground-truth probability of the $r$-th sliding window belonging to the $k$-th part of the $i$-th image, and $\hat{y}_{i}^{r(k)}$ signifies the corresponding predicted probability value.

{\bfseries Ground-truth Label of Sliding Windows}
For a given sliding window $r(u_{r}, l_{r})$ with upper and lower boundaries $u_{r}$ and $l_{r}$, respectively, we associate this window with a ground-truth label vector $ {\bf{y}}^{r}=(y^{r(1)},y^{r(2)}, ... ,y^{r(K)},y^{r(K+1)})$. The value of each element in ${\bf{y}}^{r}$ depends on the size of the overlap between $r$ and the corresponding body part or background category. In more detail, we first calculate the ground-truth upper and lower boundaries for the $k$-th part:
\begin{equation}
\begin{aligned}
u_{k} &= U + (k - 1)\times \frac{V - U}{K}, \\
l_{k} &= u_{k} + \frac{V - U}{K},
\end{aligned}
\end{equation}
where \emph{U} and \emph{V} represent the annotated upper and lower boundaries of the pedestrian in the training image. The ground-truth area for the $k$-th part is denoted as $p_{k}(u_{k}, l_{k})$. Then,
\begin{equation}
y^{r(k)} = \frac {S(p_{k}(u_{k}, l_{k}) \cap r(u_{r}, l_{r}))} {S(r(u_{r}, l_{r}))}, 1 \leq k \leq K,
\end{equation}
and
$y^{r(K+1)} = 1 - \sum_{k=1}^K y^{r(k)},$
where $S(*)$ denotes the size of the area $*$.

\subsubsection{Refined Regression of Sliding Windows}
We further promote the accuracy of the vertical alignment module by means of part-specific regression tasks. As illustrated in Fig. \ref{img2}, the \emph{K} regression tasks are constructed so that they all have the same structure. However, these tasks do not share any parameters, and each task is optimized for the detection of one specific part.

\begin{figure}
  \centerline{\includegraphics[width=0.5\textwidth]{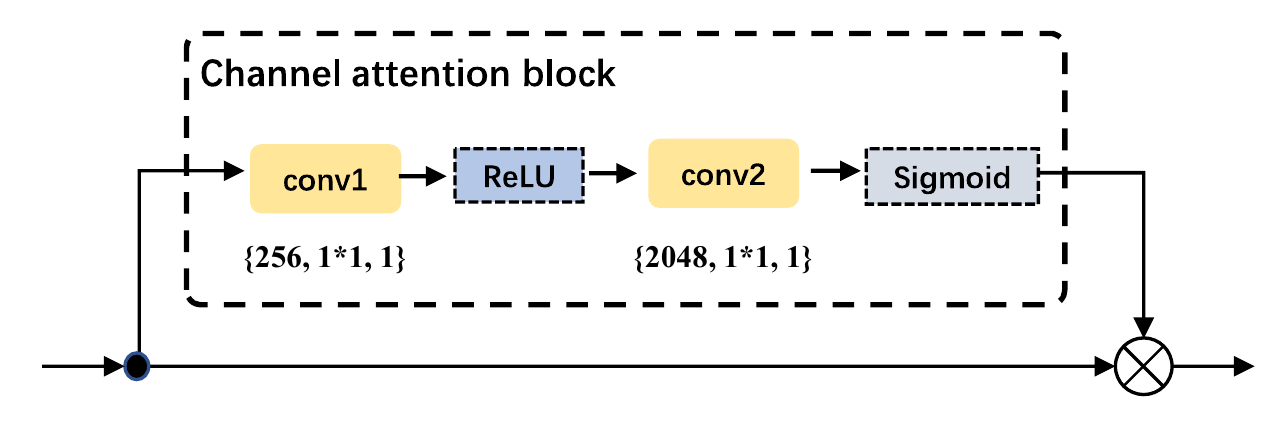}}
  \caption{Architecture of the adopted channel attention block. The three items in each bracket are: filter number, kernel size, and stride. Each convolutional layer is followed by a batch normalization layer, which is omitted in the figure in the interests of simplicity.}
  \label{img5}
\end{figure}

Each regression task incorporates one channel attention block \cite{hu2018senet}, one $1 \times 1$ convolutional layer, one FC layer, and one tanh layer. The important parameters for the above layers are labeled in Fig. \ref{img5}. The channel attention block is used to highlight the information for one specific part. Each channel attention block includes two successive $1\times 1$ convolutional operations, whose important parameters are labeled in Fig. \ref{img5}. The output of the sigmoid layer is channel attention. The input feature vector for the attention block is then multiplied with channel attention in an element-wise manner. Finally, we obtain the weighted feature vector for each sliding window.

During training, the 2048-dimensional features of all sliding windows are fed into the \emph{K} regression branches. Correspondingly, we obtain \emph{K} sets of predicted offsets after tanh normalization. Each regression task involves the optimization of the Mean Squared Error (MSE) loss:
\begin{equation}
\mathcal{L}_{r}^{k} = \frac{1}{2 \tilde{R}^{k}} \sum_{i = 1}^{N}  \sum_{j = 1}^{R}(\Delta_{j}^{i(k)} - \hat{ \Delta}_{j}^{i(k)})^2  \cdot \bm{1} \{ | \Delta_{j}^{i(k)} |  < 1 \},
\end{equation}
where $\Delta_{j}^{i(k)}$ denotes the ground-truth offset of the $j$-th sliding window for the $k$-th part in the $i$-th image, while $\hat{\Delta}_{j}^{i(k)}$ is the corresponding predicated value. $\bm{1} \{ | \Delta_{j}^{i(k)} | < 1 \}$ is equal to either 0 or 1, meaning that we only utilize sliding windows with a ground-truth offset value within the range of $(-1, 1)$. $\tilde{R}^{k}$ denotes the number of sliding windows that satisfy $\bm{1} \{ | \Delta_{j}^{i(k)} | < 1 \}$ for all $k$-th parts in a mini-batch. $\Delta_{j}^{i(k)}$ can be easily obtained by first subtracting the sliding window's center coordinate in the vertical direction from that of the $k$-th part (Eq. 5), and then normalized by the height of the sliding window. Finally, the joint loss function for the vertical alignment module can be formulated as follows:
\begin{equation}
\mathcal{L}_{v} = \mathcal{L}_{c} + \sum_{k = 1}^{K}\mathcal{L}_{r}^{k}.
\end{equation}

During testing, the $2048$-dimensional features of all sliding windows are fed into the classification task and \emph{K} regression tasks simultaneously. The classification scores and predicted offsets for the \emph{R} sliding windows are obtained for each part. We select the optimal sliding window for each part according to the following rules: first, if the classification scores of multiple sliding windows are above a pre-defined threshold \emph{T}, we select the one with the smallest offset (absolute value); second, if there is only one or no sliding window/s with a classification score above \emph{T}, we simply choose the one with the largest classification score.

\subsection{Horizontal Refinement Module}
It must be reiterated here that the above module only detects body parts in the vertical direction. Accordingly, we further propose to suppress the background information in the \emph{K} selected sliding windows via an additional horizontal refinement operation. As shown in Fig. \ref{img2}, the horizontal refinement module is applied to each part-level feature learning branch. In this paper, we realize this module using the Spatial-Channel Attention (SCA) model proposed in \cite{li2018harmonious}.

\begin{figure}
  \centerline{\includegraphics[width=0.5\textwidth]{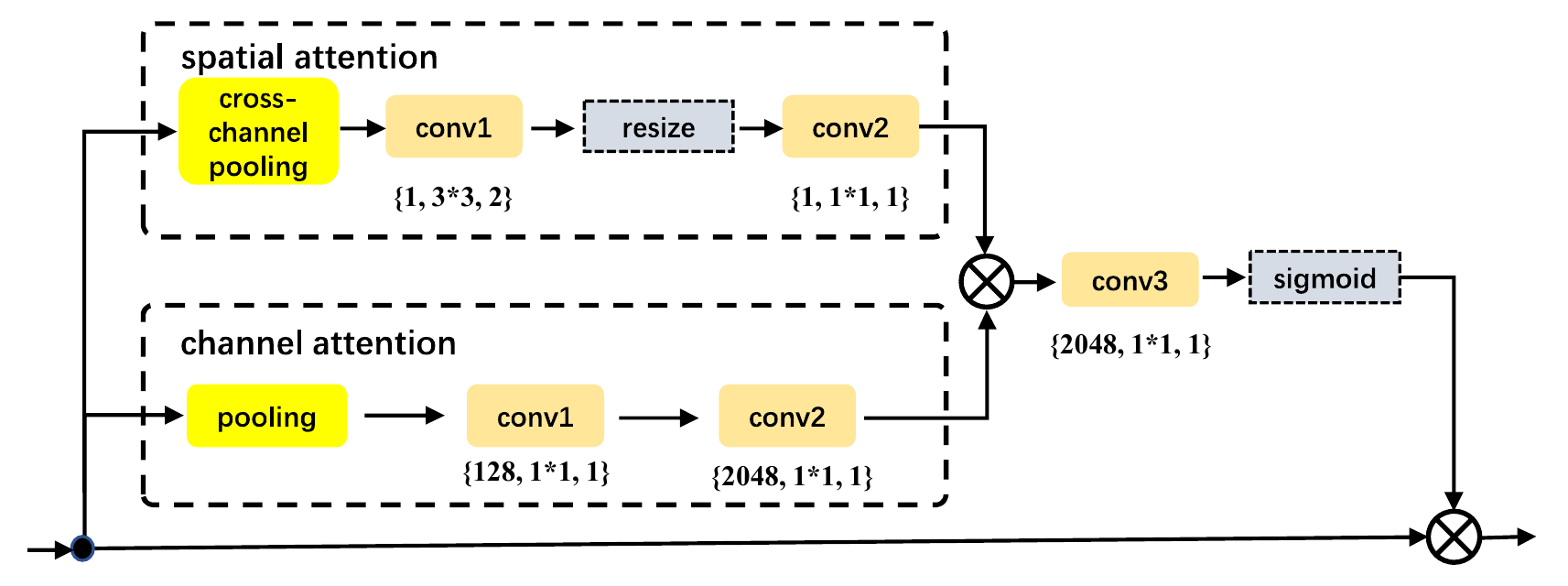}}
  \caption{Architecture of the Spatial-Channel Attention (SCA) model \cite{li2018harmonious}. We utilize SCA to realize the horizontal refinement operation. The three items in each bracket are filter number, kernel size, and stride. In the interests of brevity, the BN and ReLU layers after each convolutional layer are not shown here.
  The value of all hyper-parameters of SCA remains the same as in \cite{li2018harmonious}.}
  \label{img6}
\end{figure}

For completeness¡¯ sake, we briefly introduce the structure of SCA. As shown in Fig. \ref{img6}, the spatial and channel attentions of SCA are realized by separate branches: the former branch comprises a global cross-channel average pooling layer, a convolutional layer, a resizing bilinear layer, and another convolutional layer; the latter branch consists of a GAP layer and two successive convolutional layers. Finally, these two types of attention information are fused by one convolutional layer and normalized by one sigmoid layer. The important parameters of SCA are marked in Fig. \ref{img6}, while additional details regarding implementation can be found in \cite{li2018harmonious}.

It is worth noting here that SCA was originally employed to suppress the background information in the holistic pedestrian image \cite{li2018harmonious}, rather than around each body part. We would argue that our \emph{divide-and-conquer} strategy is more intuitive and effective, since it makes it dramatically easier to distinguish pixels of a single part from the surrounding background pixels. By comparison, applying SCA to a holistic image can be much more difficult, as both the structure of the whole body and the background information in the holistic image are often significantly more complicated.

\subsection{Person Re-ID via CDPM}
In the training stage, and taking all three modules of CDPM into account, the overall objective function for CDPM can be written as follows:
\begin{equation}
\begin{aligned}
\mathcal{L}& = \mathcal{L}_{f} + \mathcal{L}_{v} \\
 &= \sum_{k = 1}^{K}\mathcal{L}_{p}^{k} + {\lambda}_{1} * \mathcal{L}_{c} + {\lambda}_{2} * \sum_{k = 1}^{K}\mathcal{L}_{r}^{k},
\end{aligned}
\end{equation}
where ${\lambda}_{1}$ and ${\lambda}_{2}$ are the weights of the loss functions. For simplicity¡¯s sake, these are consistently set to 1 in this paper.

In the testing stage, each image passes through the backbone model to yields feature maps of size $24 \times 8 \times 2048$. For extracting the part-level features, these $24 \times 8 \times 2048$ feature maps are divided into \emph{R} sliding windows, the size of which is fixed to $4 \times 8 \times 2048$. The vertical alignment module selects one optimal sliding window for each part; subsequently, the selected sliding window for the $k$-th part passes through the $k$-th horizontal refinement module and part-level feature learning branch in order to obtain the 512-dimensional part-level feature vector ${\bf{z}}^{k}$. The final representation of the image is obtained by concatenating the above $K$ feature vectors:
\begin{equation}
{\bf{f}} = [{{\bf{z}}^{1}},{{\bf{z}}^{2}}, ..., {{\bf{z}}^{\emph{K}}}].
\end{equation}

We consistently employ the cosine distance to calculate the similarity between two image representations.

\subsection{Multi-granularity Feature}
A few recent works \cite{wang2018learning,fu2018horizontal} have adopted multi-granularity features (MGF) in order to boost ReID performance. Compared with single-level part features, MGF provides richer multi-scale information and is therefore more powerful. The proposed CDPM framework is flexible and can naturally be extended to extract MGF, which include the holistic-level feature and multi-level part features.

\begin{figure}
  \centerline{\includegraphics[width=0.50\textwidth]{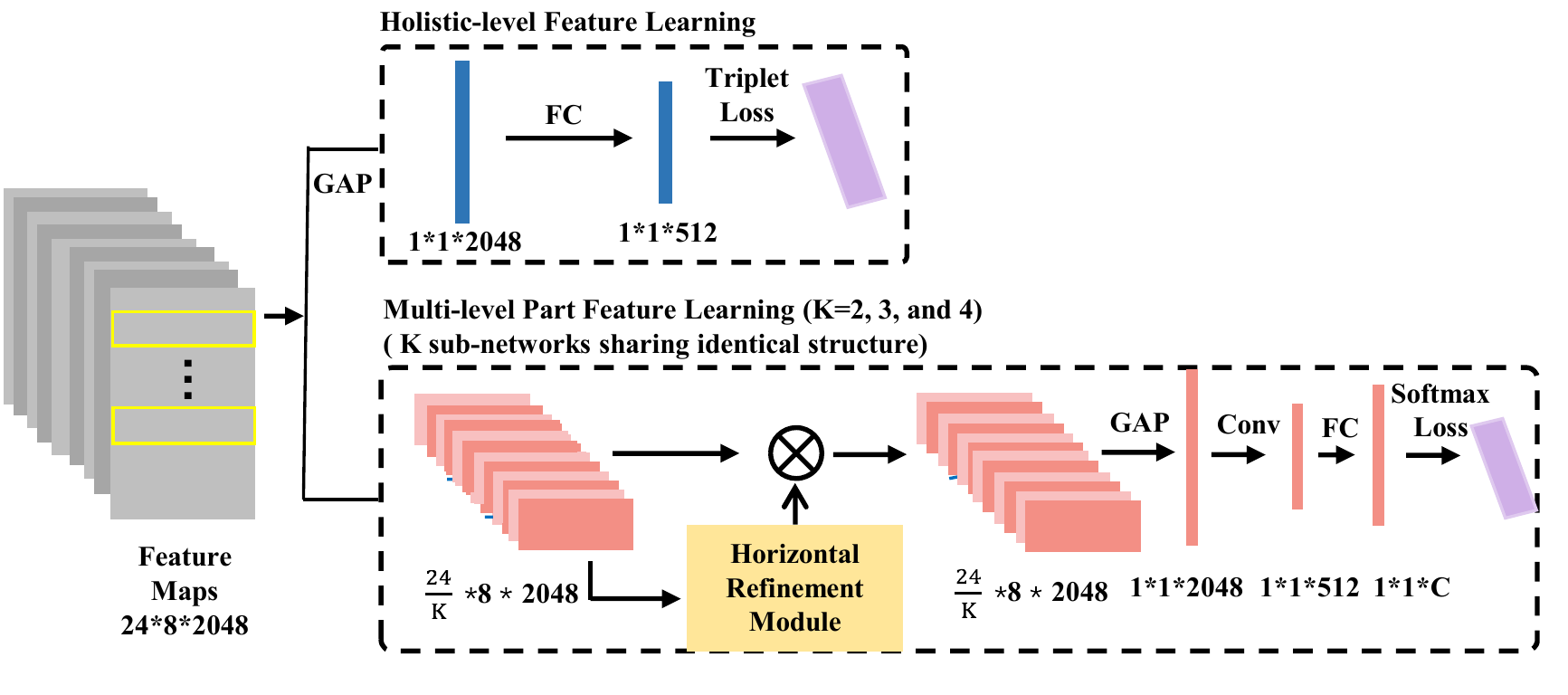}}
  \caption{New branches are equipped by CDPM to enable extraction of multi-granularity features; these include one holistic-level feature learning branch and nine additional branches for multi-level part features extraction.}
  \label{img7}
\end{figure}

\subsubsection{Holistic-level feature}
As shown in Fig. \ref{img7}, the holistic-level feature learning branch comprises one GAP layer and one FC layer. As with the part-level feature learning modules, this branch is also attached to the output of the backbone model.

Following \cite{wang2018learning}, the holistic-level feature is optimized using the triplet loss function, along with the batch-hard triplet sampling policy \cite{hermans2017defense}. To ensure that sufficient triplets are sampled in the training stage, we randomly sample \emph{A} images of each of \emph{P} random identities to compose a mini-batch. Therefore, the batch size \emph{N} is equal to $P \times A$. For each anchor image, one triplet is constructed by selecting the furthest intra-class image in the feature space as positive and the closest inter-class image as negative. The triplet loss is thus formulated as $\mathcal{L}_{g} = $
\begin{equation}
\frac{1}{2M} \sum_{i = 1}^{P} \sum_{a = 1}^{A}[\max \limits_{p=1...A} \left \| {\bf{h}}^{a}_{i} - {\bf{h}}^{p}_{i} \right \|_{2}^{2} - \min \limits_{\substack{{n=1...A}\\ {j=1...P}\\ j \neq i }} \left \| {\bf{h}}^{a}_{i} - {\bf{h}}^{n}_{j} \right \|_{2}^{2} + \alpha]_{+},
\end{equation}
where $\alpha$ denotes the margin for triplet constraint, while $M$ is the number of triplets $\left\{ {\bf{h}}^{a}_{i}, {\bf{h}}^{p}_{i}, {\bf{h}}^{n}_{j} \right\}$ in the batch that violate the constraint \cite{hermans2017defense}. Moreover, ${\bf{h}}^{a}_{i}$, ${\bf{h}}^{p}_{i}$, and ${\bf{h}}^{n}_{j}$ are L2-normalized holistic-level representations of the anchor, positive, and negative images in a triplet, respectively. $\left[*\right]_{+} = max(0, *)$ is the hinge loss.

\subsubsection{Multi-level part features}
We further add additional part-level feature learning branches of other granularities. In more detail, we set \emph{K} as 2, 3, and 4, respectively; therefore, there are nine additional part-level feature learning branches. As illustrated in Fig. \ref{img7}, both the structure and loss functions of the new branches are exactly the same as the original ones in the proposed CDPM.

\begin{figure}
  \centerline{\includegraphics[width=0.50\textwidth]{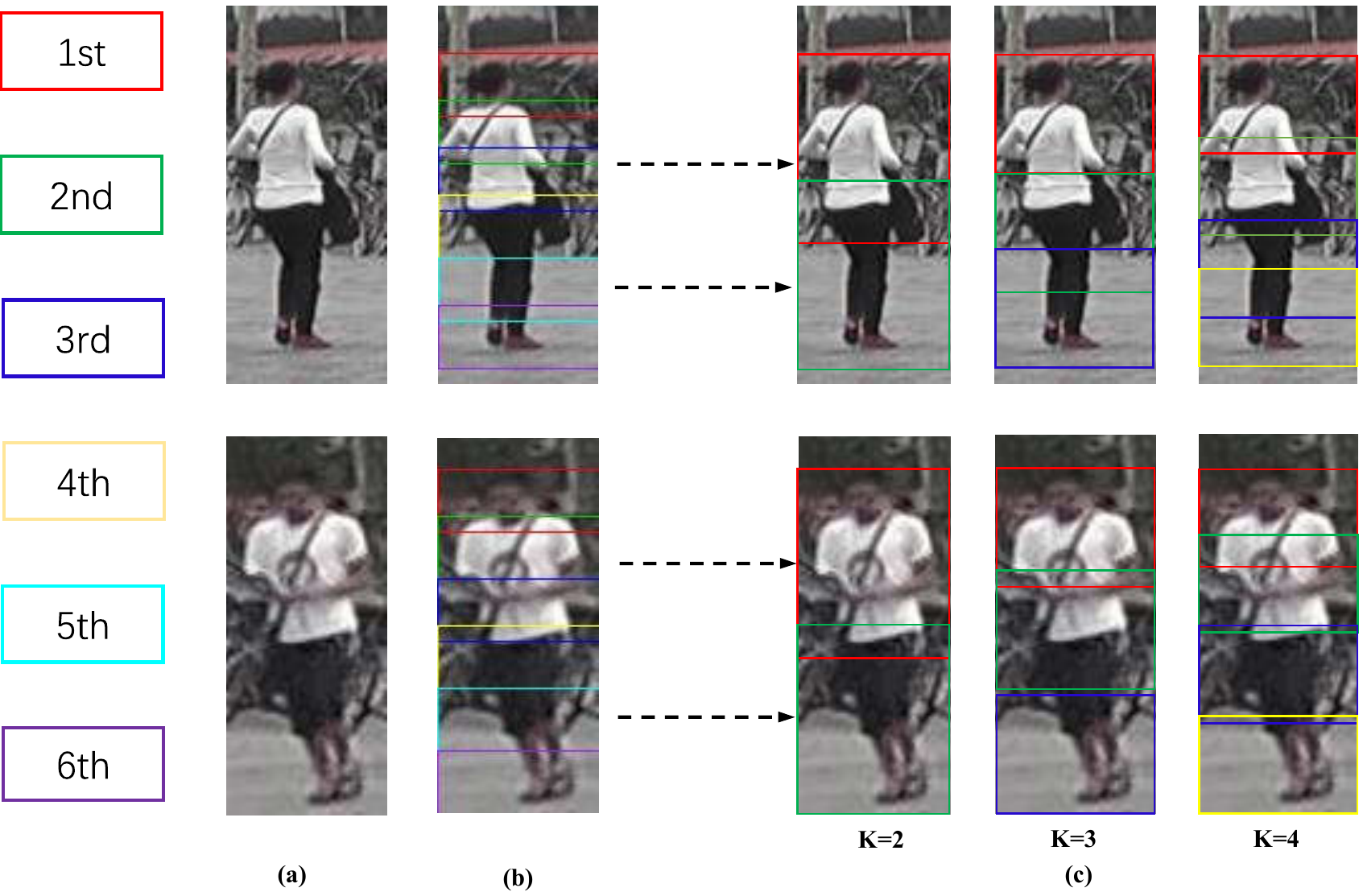}}
  \caption{Methods of obtaining the part location of new granularities in the testing stage.  (a) The original images. (b) Locations of parts of the original granularity predicted by the vertical alignment module. (c) We infer the part location of new granularities from the relevant parts in (b). For example, the center of the first part, where \emph{K} is equal to 2, can be calculated by averaging the center locations of the first three parts in (b). We fix the size of parts that have the same granularity. Best viewed in color.}
  \label{img8}
\end{figure}

It is worth noting here that the additional branches are only added in the feature learning module. The vertical alignment module of CDPM remains unchanged. As explained in Fig. \ref{img8}, in the testing stage, the location of each part of the new granularities can be inferred from the prediction results of the original vertical alignment module in CDPM.

To construct MGF in the testing stage, we extract the part-level features of all of the above granularities, and the holistic-level feature. All of the above features are concatenated to form the final representation for one pedestrian image.

\section{Experiments}

\subsection{Datasets}
To demonstrate the effectiveness of CDPM, we conduct exhaustive experiments on three large-scale person ReID benchmarks, i.e., Market-1501 \cite{Zheng2015Scalable}, DukeMTMC-ReID \cite{zheng2017unlabeled} and CUHK03 \cite{li2014deepreid}. We follow the official evaluation protocol for each database. Besides, we report both the Rank-1 accuracy and mean Average Precision (mAP) for all three datasets.

Market-1501 contains 32,668 pedestrian images. The pedestrians were detected using a DPM-based algorithm. These images depict 1,501 identities and were captured by six cameras. This dataset is divided into two sets: a training set containing 12,936 images of 751 identities, and a testing set comprising images of the remaining 750 identities. The testing set is further subdivided into a gallery set of 19,732 images and a query set of 3,368 images. We report results under both single-query and multi-query settings.

DukeMTMC-ReID features dramatic variations in both background and viewpoints. This dataset contains 36,411 images of 1,404 identities. The images were captured by eight cameras. The dataset is split into one training set containing 16,522 images of 702 identities, and one testing set comprising 17,661 gallery images and 2,228 query images of the remaining 702 identities.

CUHK03 includes 14,097 images of 1,467 identities. Images of each identity were captured by two disjoint cameras. This dataset provides both hand-labeled and DPM-detected bounding boxes. We evaluate our method using both types of bounding boxes. Besides, we adopt the new train/test protocol proposed in \cite{zhong2017re}. The new protocol splits the dataset into a training set of 767 identities and a testing set of the remaining 700 identities.

\subsection{Implementation Details}
The naive combination of the backbone model and the feature learning module, i.e., the PCB model \cite{sun2017beyond}, is selected as the baseline. Compared with CDPM, the baseline model lacks the part alignment ability. We here uniformly divide the feature maps produced by the backbone model in the vertical direction to create \emph{K} non-overlapped parts, which become the input for the part-level feature learning branches.

\subsubsection{Hyper-parameters of CDPM} The number of body parts, i.e., \emph{K}, is set to 6 following the baseline model \cite{sun2017beyond}. Besides, the threshold value \emph{T} for sliding window selection is empirically set to 0.60 for Market-1501 and 0.35 for the other two databases. The value of the hyper-parameters of SCA are kept the same as in the original work \cite{li2018harmonious}. Finally, when triplet loss is utilized for training, we set \emph{P} to 6 and \emph{A} to 8, while the margin $\alpha$ for the triplet loss is set to 0.4.

\subsubsection{Training details}
Experiments are conducted using the PyTorch framework. All pedestrian images are resized to $384 \times 128$ pixels. In line with existing works \cite{li2014deepreid, zhong2017random}, we adopt extensive data augmentation to reduce overfitting. Specifically, we augment the training data via offline translation \cite{li2014deepreid}, online horizontal flipping, and online random erasing \cite{zhong2017random}. After the offline translation, the size of each training set is enlarged by 5 times. Moreover, the ratio of random erasing is set to 0.5.

 We use the standard stochastic gradient descent with momentum \cite{sutskever2013importance} for model optimization and further set the momentum value to 0.9 and the batch size \emph{N} to 48. Besides, we utilize a stage-wise strategy to train the proposed CDPM. In the first stage, we fine-tune the baseline model from the IDE model proposed in \cite{zheng2017person} for 50 epochs; here, the learning rate is initially set to 0.01 and then multiplied by 0.1 for every 20 epochs. In the second stage, we fix the parameters of the baseline model and optimize only the parameters of the components newly introduced in CDPM (i.e., the vertical alignment module and horizontal refinement modules). This stage is trained for 40 epochs, with the learning rate set to 0.01 initially and then multiplied by 0.1 for every 15 epochs. Finally, all CDPM model parameters are fine-tuned in an end-to-end fashion for 30 epochs, with a small initial learning rate of 0.001 that decreases to 0.0001 after 20 epochs.

\subsection{Comparisons to State-of-the-Art Methods}
The essential contribution of CDPM lies in its ability to detect flexible body parts for ReID.
For fair comparison with existing approaches, we categorize them into three groups, i.e., holistic feature-based methods, single-level part feature-based methods, and multi-granularity feature-based methods. Hereafter, they are denoted as `HF-based', `SPF-based', and `MGF', respectively.

\subsubsection{Evaluation on Market-1501 Dataset}
Performance comparisons between CDPM and state-of-the-art methods on the Market-1501 database are tabulated in Table \ref{tbl:table1}. From the table, it can be seen that CDPM significantly outperforms all existing methods for both Rank-1 and mAP results, and achieves state-of-the-art performance. In particular, CDPM outperforms the most recent SPF-based method, i.e., PCB+RPP \cite{sun2017beyond}, by {{1.4\% and 4.4\%}} on the single-query mode for Rank-1 accuracy and mAP, respectively. The above comparisons successfully demonstrate the effectiveness of CDPM. Moreover, the performance of CDPM is further improved through the extraction of multi-granularity features. Finally, CDPM* achieves state-of-the-art performance superior to that of all other approaches. Specifically, CDPM* achieves {{95.9\% and 87.2\%}} for Rank-1 accuracy and mAP on the single-query mode, respectively.

\begin{table}[h]
\caption{Performance Comparisons on the Market-1501 Dataset. Both Rank-1 Accuracy (\%) and mAP (\%) Indices Are Compared. CDPM* Refers to the CDPM Model that Extracts Multi-granularity Features for ReID}
\centering
\begin{tabular}{|c|c|cc|cc|}
\hline
\multicolumn{2}{|c|}{Query Type} & \multicolumn{2}{c|}{Single Query} & \multicolumn{2}{c|}{Multiple Query} \\
\hline
\multicolumn{2}{|c|}{Methods} & Rank-1  & mAP  & Rank-1 & mAP \\
\hline
 \multirow{6}*{\rotatebox{90}{HF-based}}
    &SVDNet \cite{sun2017svdnet} &82.3 &62.1 & - & - \\
    &PAN \cite{zheng2018pedestrian} &82.8 &63.4& 88.2 & 71.7 \\
    &MGCAM \cite{song2018mask} &83.6 & 74.3 & - & - \\
    &Triplet Loss \cite{hermans2017defense} &84.9 &69.1 &90.5 &76.4 \\
    &DaRe \cite{wang2018resource} &86.4 &69.3 &- &- \\
    &MLFN \cite{chang2018multi} & 90.0 &74.3 &92.3 & 82.4 \\
\hline
 \multirow{9}*{\rotatebox{90}{SPF-based}}
 &Spindle \cite{zhao2017spindle} &76.9 & - & - & - \\
 &MSCAN \cite{li2017learning} &80.3 & 57.5 & 86.8 & 66.7 \\
 &PAR \cite{zhao2017deeply} &81.0 &63.4 & - & - \\
 &PDC \cite{su2017pose} & 84.1 &63.4 & - & - \\
 &AACN \cite{xu2018attention} & 85.9 &66.9 & 89.8 & 75.1 \\
 &HA-CNN \cite{li2018harmonious} &91.2 &75.7 &93.8 &82.8 \\
 &AlignedReID \cite{zhang2017alignedreid} &91.8 &79.3 &- & - \\
 &PCB+RPP \cite{sun2017beyond} & 93.8 &81.6 & - & - \\
 &CDPM &{{\bfseries 95.2}} &{{\bfseries 86.0}} &{{\bfseries 96.4}}&{{\bfseries 89.9}} \\
\hline
 \multirow{3}*{\rotatebox{90}{MGF}}
 &HPM \cite{fu2018horizontal} & 94.2 &82.7 & - & - \\
 &MGN \cite{wang2018learning} &95.7 &86.9 &96.9 &90.7 \\
 &CDPM* &{{\bfseries 95.9}} &{{\bfseries 87.2}} &{{\bfseries 97.0}}&{{\bfseries 91.1}} \\
\hline
\end{tabular}
\label{tbl:table1}
\end{table}

\subsubsection{Evaluation on DukeMTMC-ReID Dataset}
Compared with Market-1501, the pedestrian images in the DukeMTMC-ReID database are impacted by more variations in viewpoint and background. Performance comparisons are summarized in Table \ref{tbl:table2}. These results reveal that CDPM achieves the best Rank-1 accuracy and mAP overall, outperforming all other state-of-the-art methods by a large margin. In particular, CDPM outperforms the best existing SPF-based approach \cite{suh2018part} by 3.8\% and 8.2\% for Rank-1 accuracy and mAP, respectively. These results suggest that CDPM can locate body parts accurately, even in cases where dramatic variations exist in terms of the viewpoints and background.

\begin{table}
\centering\caption{Performance Comparisons on DukeMTMC-ReID Dataset. CDPM* Refers to the CDPM Model that Extracts Multi-granularity Features for ReID}
\centering
\begin{tabular}{|c|c|cc|}
\hline
  \multicolumn{2}{|c|}{Methods} & Rank-1 & mAP \\
\hline
 \multirow{4}*{\rotatebox{90}{HF-based}}
  & PAN \cite{zheng2018pedestrian} &71.6 & 51.5\\
  & DaRe \cite{wang2018resource} &75.2 & 57.4\\
  & SVDNet \cite{sun2017svdnet} &76.7 & 56.8\\
  & MLFN \cite{chang2018multi} &81.0 &62.8 \\

\hline
  \multirow{5}*{\rotatebox{90}{SPF-based}}
  & AACN \cite{xu2018attention} & 76.8 & 59.3\\
  & HA-CNN \cite{li2018harmonious} &80.5 & 63.8\\
  & PCB+RPP \cite{sun2017beyond} & 83.3 & 69.2 \\
  & Part-aligned \cite{suh2018part} &84.4 & 69.3\\
  & CDPM &{{\bfseries 88.2}} &{{\bfseries 77.5}} \\
\hline
  \multirow{3}*{\rotatebox{90}{MGF}}
  & HPM \cite{fu2018horizontal} & 86.6 & 74.3 \\
  & MGN \cite{wang2018learning} & 88.7 &78.4 \\
  & CDPM*  &{{\bfseries 90.1}} &{{\bfseries 80.2}} \\
  \hline
\end{tabular}
\label{tbl:table2}
\end{table}

\subsubsection{Evaluation on CUHK03 Dataset}
We evaluate the performance of CDPM on CUHK03 using both manually labeled and auto-detected bounding boxes. The results of the comparison are tabulated in Table \ref{tbl:table3}. From the table, it can be seen that CDPM achieves the best performance among all SPF-based approaches. In particular, it outperforms the second-best approach \cite{sun2017beyond} by {{8.2\% and 9.5\%}} for Rank-1 accuracy and mAP using auto-detected bounding boxes, respectively. Furthermore, CDPM* also achieves the best performance among MGF-based approaches. These above comparisons clearly justify the overall effectiveness of CDPM.

\begin{table}
\centering
\caption{Performance Comparisons on the CUHK03 Dataset, Using the New Protocol Proposed in \cite{zhong2017re}. CDPM* Refers to the CDPM Model that Extracts Multi-granularity Features for ReID}
\begin{tabular}{|c|c|cc|cc|}
\hline
  \multicolumn{2}{|c|} {Bounding Boxes Type} & \multicolumn{2}{c|}{detected} & \multicolumn{2}{c|}{labeled} \\
\hline
  \multicolumn{2}{|c|}{Methods} & Rank-1  & mAP  & Rank-1 & mAP \\
\hline
  \multirow{6}*{\rotatebox{90}{HF-based}}
  & PAN \cite{wang2018resource} &36.3 &34.0 &36.9 &35.0 \\
  & SVDNet \cite{sun2017svdnet} &41.5 &37.3 & - & - \\
  & DPFL \cite{chen2017person} &43.0 &40.5 &40.7 & 37.0 \\
  & MGCAM \cite{song2018mask} &46.7 & 46.9 & 50.1 & 50.2 \\
  & MLFN \cite{chang2018multi} &52.8 &47.8 &54.7 &49.2 \\
  & DaRe \cite{wang2018resource} & 55.1 &51.3 &58.1 &53.7 \\
\hline
  \multirow{3}*{\rotatebox{90}{SPF-based}}
  & HA-CNN \cite{li2018harmonious} &41.7 &38.6 &44.4 &41.0 \\
  & PCB+RPP \cite{sun2017beyond} & 63.7 &57.5 & - & - \\
  & HPDN \cite{zhang2018person62} &- &- &64.3 &58.2 \\
  & CDPM &{{\bfseries 71.9}} &{{\bfseries 67.0}} &{{\bfseries 75.8}}&{{\bfseries 71.1}} \\
\hline
 \multirow{3}*{\rotatebox{90}{MGF}}
  & HPM \cite{fu2018horizontal} & 63.9 &57.5 & - & - \\
  & MGN \cite{wang2018learning} &66.8 &66.0 &68.0 &67.4 \\
  & CDPM* &{{\bfseries 78.8}} &{{\bfseries 73.3}} &{{\bfseries 81.4}}&{{\bfseries 77.5}} \\
  \hline
\end{tabular}
\label{tbl:table3}
\end{table}

\subsection{Ablation Study}
In the following, we present the results of ablation study conducted to justify the effectiveness of each newly introduced component of CDPM, i.e., the vertical alignment module and horizontal refinement module. Moreover, we also compare some possible variants of the vertical alignment module and design several experiments to evaluate the influence of the annotation on the performance of CDPM. At last, efficiency comparison and some visualizations are performed to further verify the effectiveness and superiority of CDPM. In line with recent works \cite{li2018harmonious,sun2017beyond}, the ablation study is conducted on both the Market-1501 and DukeMTMC-ReID datasets.

\subsubsection{Effectiveness of the Vertical Alignment Module}
\begin{figure}
  \centerline{\includegraphics[width=0.50\textwidth]{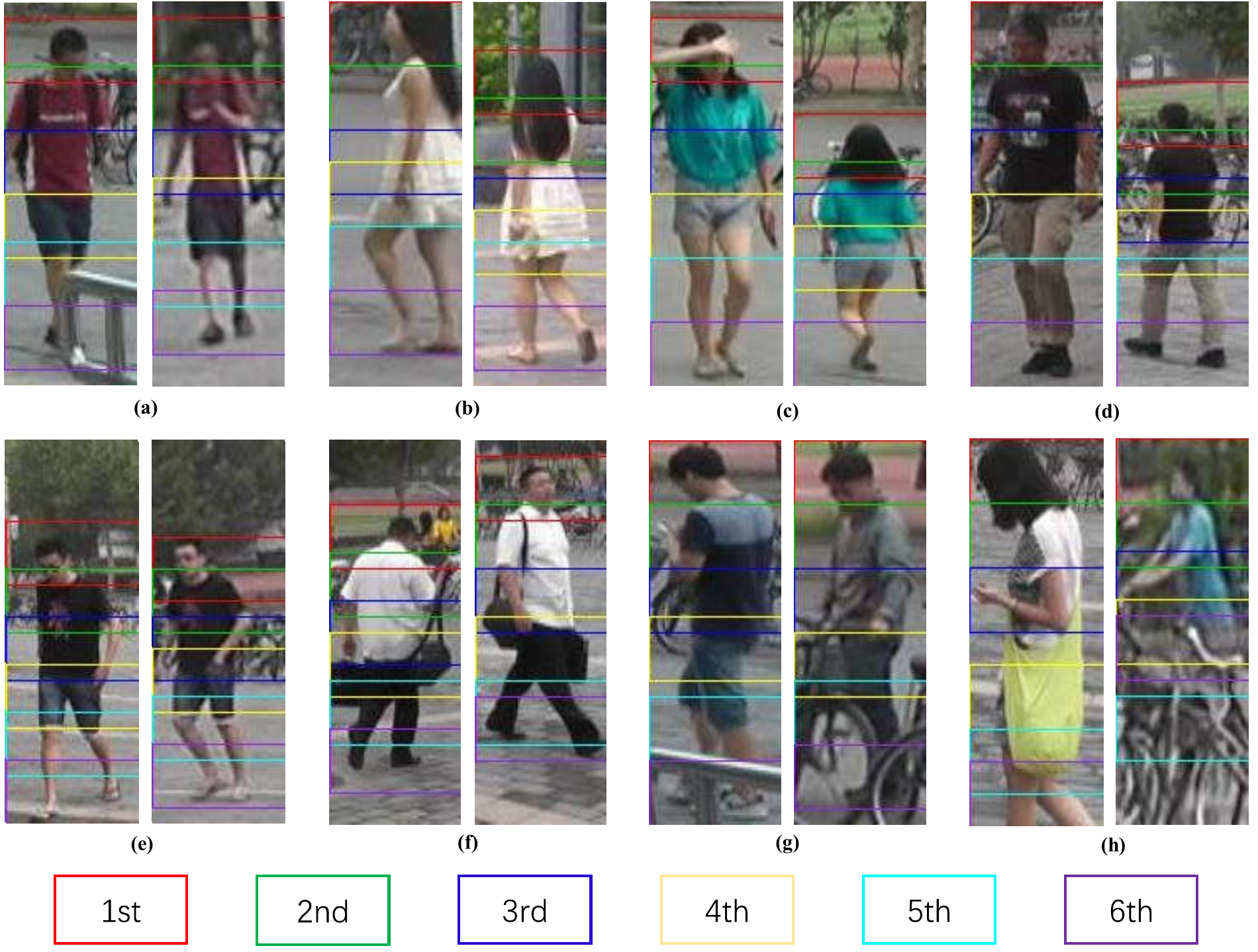}}
  \caption{The vertical alignment module can detect body parts robustly under a vast majority of circumstances, including moderate (a, b) or even severe (c, d) part misalignment, dramatic pose variations (e, f), and occlusion (g). However, the vertical alignment module may fail in cases where complex occlusion or a part missing problem exist (h). Best viewed in color.}
  \label{img9}
\end{figure}

In this subsection, we equip the baseline model with the vertical alignment module and denote this model as Baseline+V in Table \ref{tbl:table4}. As shown in Table \ref{tbl:table4}, equipping the vertical alignment module consistently promotes ReID performance. In particular, Baseline+V outperforms the baseline model by 1.1\% and 1.5\% in terms of Rank-1 accuracy on Market-1501 and DukeMTMC-ReID, respectively. Moreover, exemplars of body-part detection results achieved by the vertical alignment module are presented in Fig. \ref{img9}. It can be clearly seen from the figure that the vertical alignment module is able to detect body parts rather robustly even in cases of severe misalignment, occlusion, and pose variations. We can also observe that the vertical alignment module may fail in the event of complex occlusion or a part missing problem. The above experiments justify the effectiveness of the vertical alignment module.

Moreover, to further verify the superiority of the proposed part-based vertical alignment module, we also evaluate a variant that detects only the upper and lower boundaries of the pedestrian in one image. This model is realized by keeping the two regression tasks for the first and the last body parts while removing the other \emph{K}-2 regression tasks; we refer to this model as Baseline+V(G) in Table IV. In the testing stage, we define the upper and lower boundaries of one pedestrian as the upper boundary of the first part and the lower boundary of the last part, respectively. After the two boundaries of one pedestrian have been predicted using the two regression tasks, the feature maps are uniformly divided in the vertical direction between the two boundaries in order to produce the location for each body part.

The experimental results tabulated in Table IV clearly demonstrate that Baseline+V significantly outperforms Baseline+V(G). For example, Baseline+V obtains 94.6\% in Rank-1 accuracy and 84.5\% in mAP on the Market-1501 database, surpassing Baseline+V(G) by 0.9\% in Rank-1 accuracy and 2.3\% in mAP, respectively. The main reason is Baseline+V(G) is less reliable in body part detection. Detection errors in either of the two boundaries will cause errors in the location of all body parts. In comparison, the detection of \emph{K} body parts is independent in Baseline+V.  Therefore, Baseline+V is more robust in body part detection. The above results clearly demonstrate the superiority of the proposed part-based detection scheme of CDPM.

\subsubsection{Effectiveness of the Horizontal Refinement Module}
In this subsection of the experiments, we equip the baseline model with the horizontal refinement module only, which is denoted as Baseline+H in Table \ref{tbl:table4}. It is worth recalling here that Baseline+H applies one SCA \cite{li2018harmonious} module to each part-level feature learning branch. As can be seen from Table \ref{tbl:table4}, the horizontal refinement module achieves consistently superior results to those of baseline. For example, it improves the Rank-1 accuracy on Market-1501 from 93.5\% to 94.7\%, which represents a 18.5\% relative reduction in the error rate.

Moreover, similar to \cite{li2018harmonious}, we also try to apply only one SCA module to the whole feature maps output by the backbone model; this approach is denoted as Baseline+H(G) in Table \ref{tbl:table4}. Experimental results indicate that Baseline+H consistently outperforms Baseline+H(G). This result justifies our motivation that it is much easier to reduce the interference of background information in a \emph{divide-and-conquer} manner.

Finally, we simultaneously equip the baseline model with both modules. The results can be found in the row `CDPM' in Table \ref{tbl:table4}. From these results, it is clear that the combination of the two modules creates a considerable performance boost relative to the use of one module. Compared with the baseline model, CDPM boosts the Rank-1 accuracy by {{1.7\% and 2.2\%}}, and mAP by {{4.1\% and 3.0\%}}, on the Market-1501 and DukeMTMC-ReID datasets, respectively. From the above comparisons, we conclude that the vertical alignment module and horizontal refinement module are complementary; therefore the proposed \emph{divide-and-conquer} solution is effective.

\begin{table}
\caption{Evaluation of the Effectiveness of Each Component in CDPM. Here, V Denotes the Vertical Alignment Module and H Denotes the Horizontal Refinement Module\protect\footnotemark[1]}
\centering
\begin{tabular}{|p{2.5cm}<{\centering}|cc|cc|}
\hline
  Dataset & \multicolumn{2}{c|}{Market-1501} & \multicolumn{2}{c|}{DukeMTMC-ReID} \\
  \hline
  Metric  & Rank-1  & mAP  & Rank-1 & mAP \\
  \hline
  Baseline &93.5 &81.9 &86.0 &74.5 \\
  \hline
  Baseline+V  &{94.6} &{84.5} &{87.5} &{76.1}   \\
  Baseline+V(G) &{93.7} &{82.2}  &{86.3}  &{74.9}  \\
  \hline
  Baseline+H  &94.7 &84.8 &87.7 &76.7   \\
  Baseline+H(G) &94.0 &83.6 &86.5 &75.4     \\
  \hline
  CDPM &{95.2} &{86.0}  &{88.2}  &{77.5} \\
  \hline
\end{tabular}
\label{tbl:table4}
\end{table}
\footnotetext[1]{For fair comparison, all models in Table \ref{tbl:table4} are trained with the same number of epochs.}

\subsubsection{Structure of the Vertical Alignment Module}
In this subsection, we compare the performance of the vertical alignment module with two possible variants. The first variant (denoted as Variant 1 in Table \ref{tbl:table5}), which is similar to Faster-RCNN \cite{ren2017faster}, shares the parameters of the $1 \times 1$ convolutional layer of the coarse classification task and \emph{K} refined regression tasks. The second variant (denoted as Variant 2 in Table \ref{tbl:table5}) shares the parameters of the $1 \times 1$ convolutional layer between each part-level feature learning branch and the corresponding refined regression task in the vertical alignment module. As can be seen from Table \ref{tbl:table5}, both of these variants are inferior to the proposed CDPM. For example, the Rank-1 accuracies of the two variants on the Market-1501 database are lower than those achieved by CDPM by {{0.9\% and 3.0\%}}, respectively.

Two insights can be gained from the above results: firstly, in contrast with object detection \cite{ren2017faster}, body part detection is a fine-grained task, meaning that more independent parameters are required for each specific part detection task; secondly, part-level feature learning for recognition and body part detection are heterogenous tasks, in that the former learns the unique characteristic of one identity while the latter is based on the general characteristic of one specific body part between different identities. Therefore, these two tasks should not share parameters.

\begin{figure}
  \centerline{\includegraphics[width=0.50\textwidth]{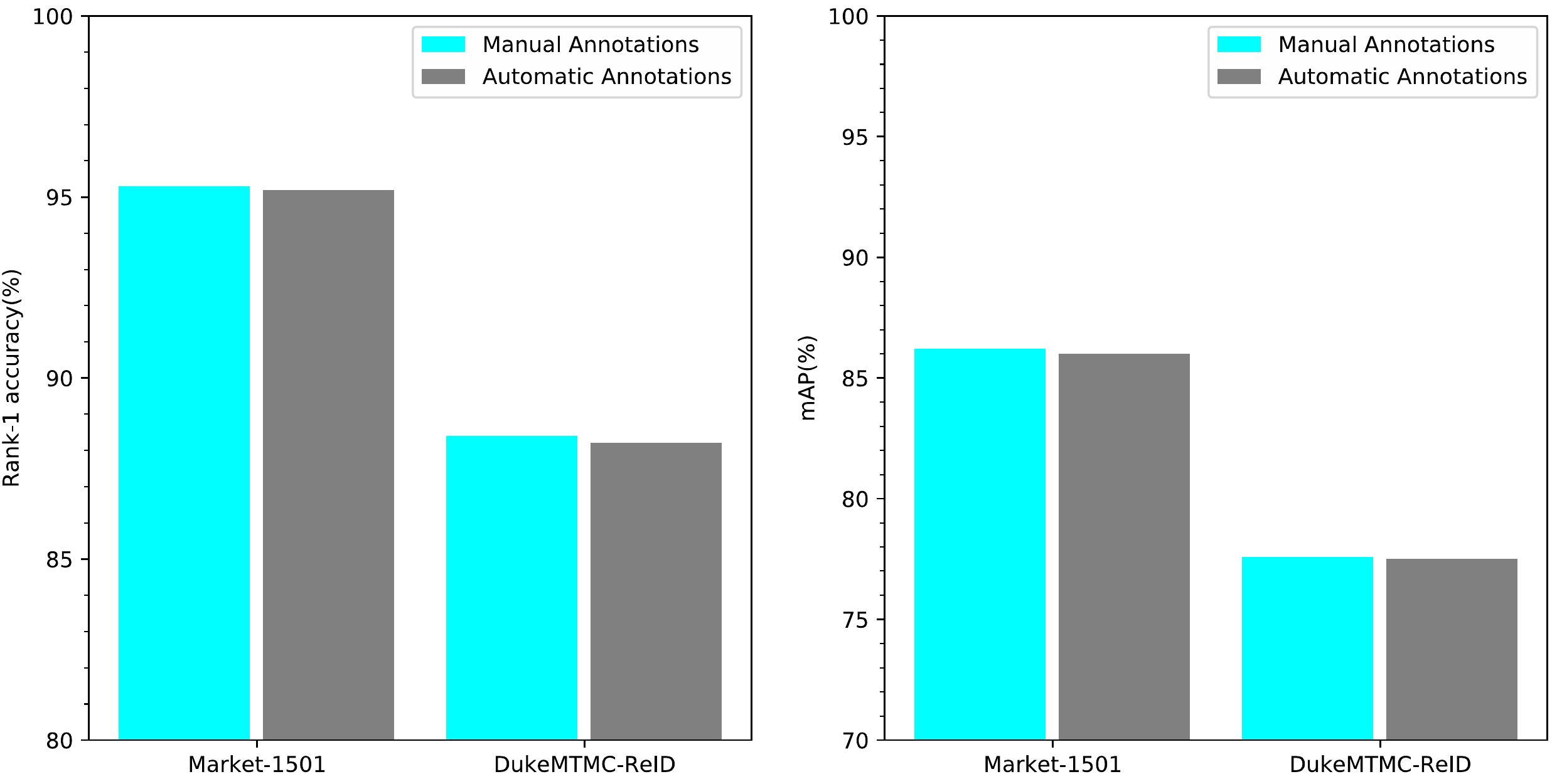}}
  {\caption{Performance comparison of CDPM with automatic annotations and manual annotations.}
  \label{img10}
  }
\end{figure}

\begin{table}
\caption{Performance Comparisons of Variants for the Vertical Alignment Module}
\centering
\begin{tabular}{|c|cc|cc|}
\hline
  Dataset & \multicolumn{2}{c|}{Market-1501} & \multicolumn{2}{c|}{DukeMTMC-ReID} \\
  \hline
  Metric  & Rank-1  & mAP  & Rank-1 & mAP \\
  \hline
  Baseline &93.5 &81.9 &86.0 &74.5 \\
  \hline
  Variant 1 &{94.3} &{85.4} &{87.4} &{76.0} \\
  Variant 2 &{92.2} &{80.7} &{84.9} &{73.2} \\
  \hline
  CDPM &{95.2} &{86.0}  &{88.2}  &{77.5}  \\
  \hline
\end{tabular}
\label{tbl:table5}
\end{table}

\subsubsection{Comparisons in Different Types of Annotations}
In the following, we explore the influence of the quality of annotations by comparing the performance of the proposed CDPM when trained with automatic annotations as opposed to manual annotations.

From Fig. \ref{img10}, it is clear that the performance of CDPM on both the Market-1501 and DukeMTMC-ReID databases are fairly similar regardless of which type of annotations is used. For example, compared with manual annotations, the performance of CDPM with automatic annotations is reduced by only 0.1\% in Rank-1 accuracy and 0.2\% in mAP on the Market-1501 database. These results demonstrate that CDPM is robust to the quality of annotations, meaning that it will be easy to use in real-world applications.

\subsubsection{Efficiency Comparison}
We compare the efficiency of CDPM with some state-of-the-art works \cite{zhao2017spindle,su2017pose,sun2017beyond,wang2018learning}, then tabulate the results of this comparison in Table VI. It is worth noting here that the existing works are quite different as regards the details of their implementation. Therefore, for fair comparison, we mainly draw comparisons with approaches that adopt similar backbone models. We also resize the input image for all models in Table VI so that the image size is as close to $384 \times 128$ pixels as possible.

It is also worth noting here that neither PCB \cite{sun2017beyond} or MGN \cite{wang2018learning} require part detection; they are therefore faster. In comparison, Spindle \cite{zhao2017spindle}, PDC \cite{su2017pose}, and CDPM perform part detection and part-level feature extraction separately; therefore, they require more computational costs. As shown in Table VI, CDPM is competitive in terms of efficiency when compared with Spindle \cite{zhao2017spindle} and PDC \cite{su2017pose}. For example, the time cost of CDPM in the testing stage is only about 56.7\% of PDC's \cite{su2017pose}. This is because CDPM integrates part detection and part-level feature extraction into one compact framework.

\begin{figure}
  \centerline{\includegraphics[width=0.46\textwidth]{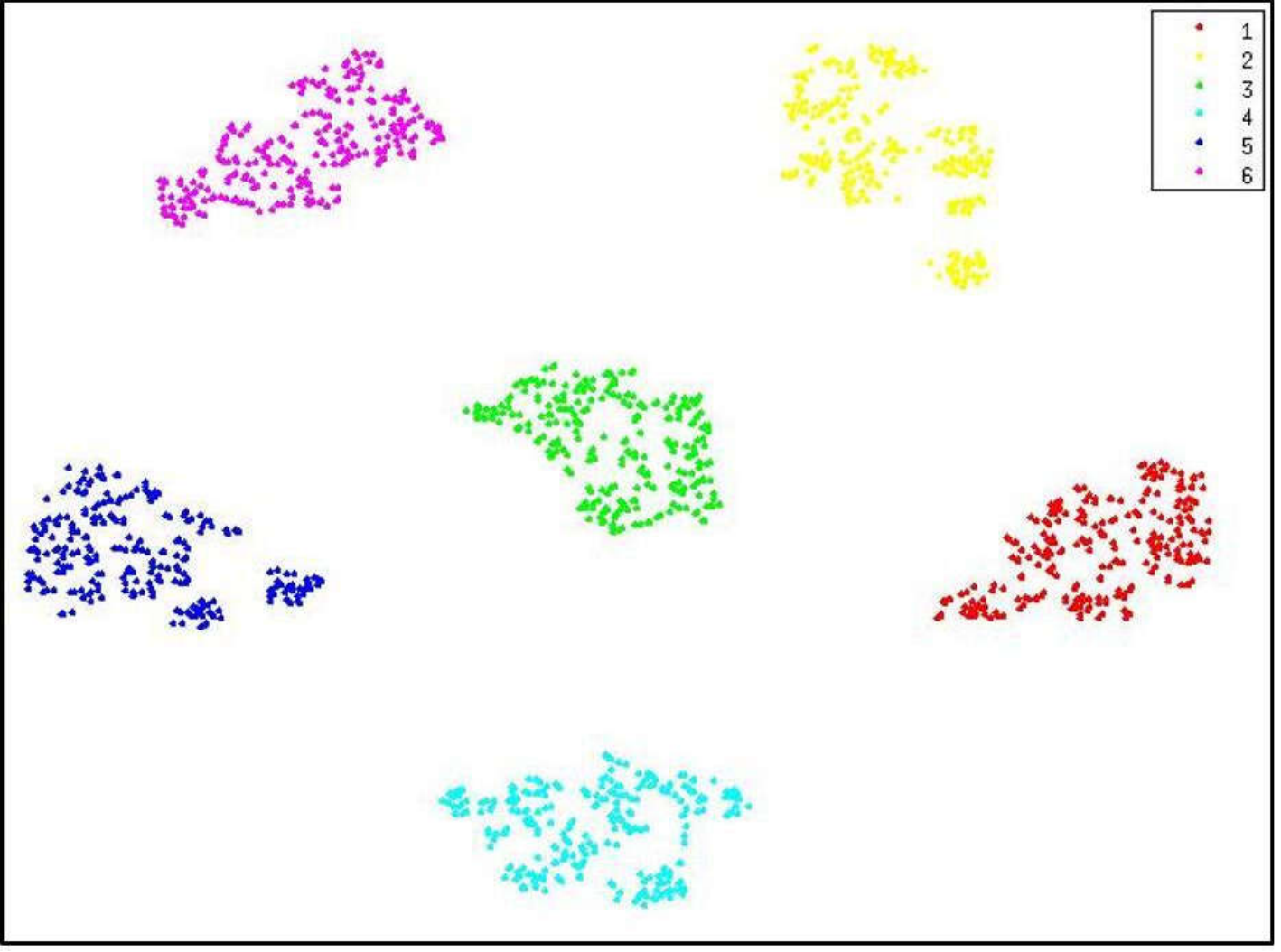}}
  {\caption{Visualization of features obtained from the last convolutional layer of each of the \emph{K} regression tasks using t-SNE \cite{maaten2008visualizing}. A total of 250 images are selected from 50 identities. \emph{K} body parts are denoted using different colors.}
  \label{img11}
  }
\end{figure}

\newcommand{\tabincell}[2]{\begin{tabular}{@{}#1@{}}#2\end{tabular}}
\begin{table}
	\centering
	{\caption{Comparisons in Efficiency of Different Models for a Single Image}
	\begin{tabular}{|p{1.42cm}<{\centering}|c|p{0.91cm}<{\centering}|p{0.91cm}<{\centering}p{0.91cm}<{\centering}c|}
		\hline
		\multirow{3}*{\tabincell{c}{Method}} &\multirow{3}*{\tabincell{c}{Backbone}}   &\multirow{3}*{\tabincell{c}{Train(ms)}}   &\multicolumn{3}{c|}{Test(ms)}    \\ \cline{4-6}
		& & &\multirow{2}*{\tabincell{c} {Part \\ Detection}}    &\multirow{2}*{\tabincell{c}{Feature \\ Extraction}}  &\multirow{2}{*}{Total} \\
        & & & & & \\
        \hline
		PCB \cite{sun2017beyond} &ResNet50        &11.7 & -   &9.4  &9.4 \\
		MGN \cite{wang2018learning} &ResNet50     &19.3 & -   &16.9 &16.9\\
        \hline
		Spindle\cite{zhao2017spindle}  &Self-Designed  &47.7 &26.7 &17.9 &44.6\\
		PDC\cite{su2017pose} &GoogleNet           &70.1 &45.6 &14.7 &60.3\\
        \hline
        CDPM &ResNet50                            &39.4 &24.6 &9.6  &34.2\\
        \hline
	\end{tabular}
    }
\label{table6}
\end{table}

\begin{figure}
  \centerline{\includegraphics[width=0.5\textwidth]{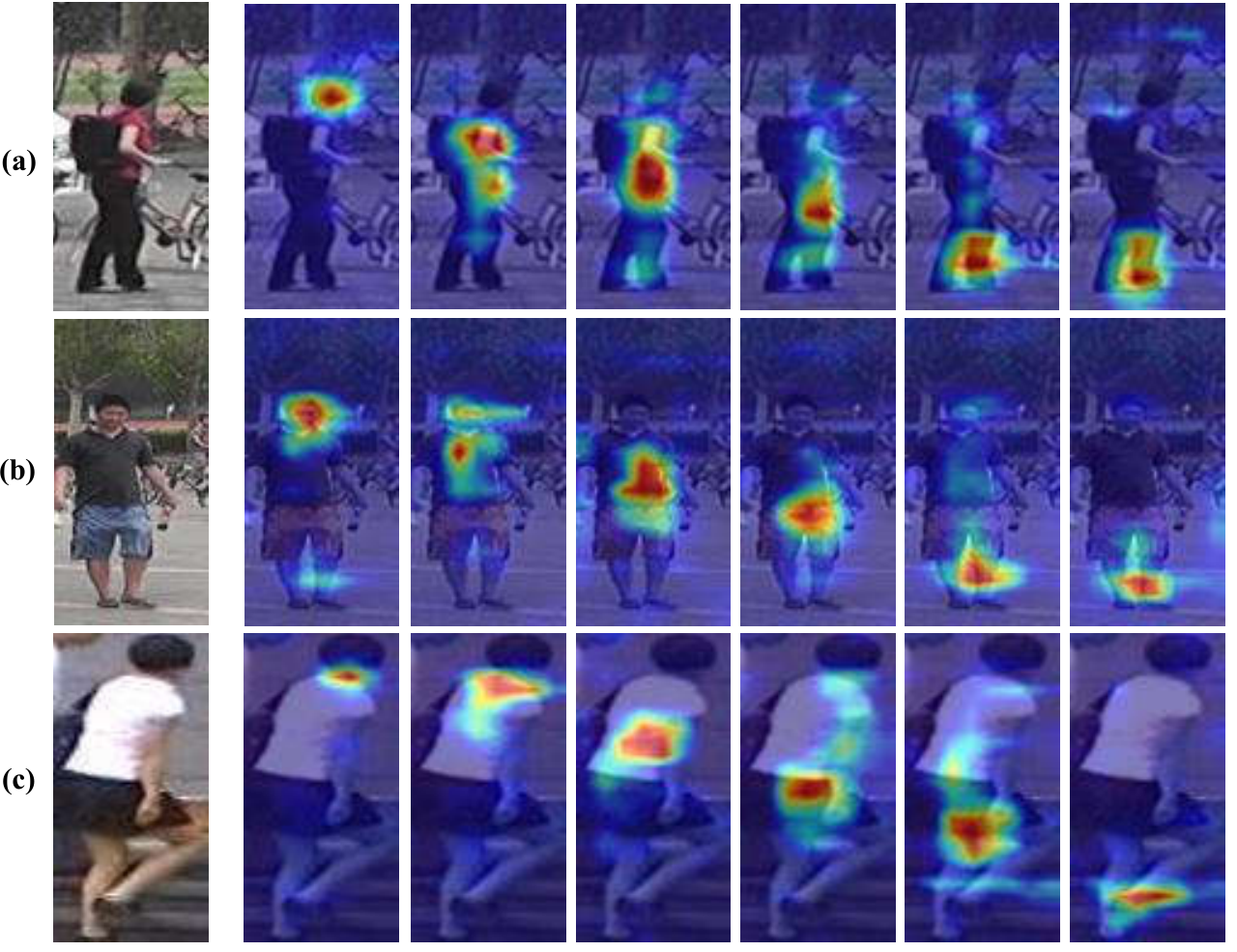}}
  {\caption{Visualization of the part-relevant saliency maps generated by each feature learning module of CDPM utilizing Grad-CAM \cite{selvaraju2017grad} for images from Market-1501. For each image, we list the generated saliency maps for the six body parts. Best viewed in color.}
  \label{img12}
  }
\end{figure}

\subsubsection{Visualization of Features for the Vertical Alignment Module}
To explain the effectiveness of the vertical alignment module, we visualize the features learned by vertical alignment module utilizing t-SNE \cite{maaten2008visualizing}. These features are extracted from the last convolutional layer of each of the \emph{K} regression tasks in the vertical alignment module for each body part. We select five images in each of 50 identities from the gallery set of the Market-1501 database. These selected images are affected to greater or lesser extents by the part misalignment problem.

From Fig. \ref{img11}, we can also clearly see that the features from the same body part are close to each other, despite being from different identities. Moreover, we also observe that the inter-part difference is significant when compared with intra-part variances. These results verify that the vertical alignment module is able to capture the general characteristic of each body part.

\subsubsection{Visualization of Part-relevant Saliency Maps}
To arrive at an interpretation of how the proposed CDPM captures the body parts, we opt to visualize the part-relevant saliency maps generated by each feature learning module of CDPM utilizing Grad-CAM \cite{selvaraju2017grad}.

The resulting saliency maps are presented in Fig. \ref{img12}. We can also clearly observe from the figure that CDPM adaptively focuses on discriminative body parts, even when these the body parts are misaligned due to of detection error (Fig. \ref{img12}(a)(b)) or variation of pose (Fig. \ref{img12}(c)). These visualization results verify the effectiveness of the proposed CDPM in solving the part misalignment problem.

\subsubsection{Visualization of Retrieval Results}
Finally, we visualize the retrieval results of CDPM and two other state-of-the-art approaches, namely PCB \cite{sun2017beyond} and MGN \cite{wang2018learning}.

\begin{figure*}
  {
  \centerline{\includegraphics[width=1.0\textwidth]{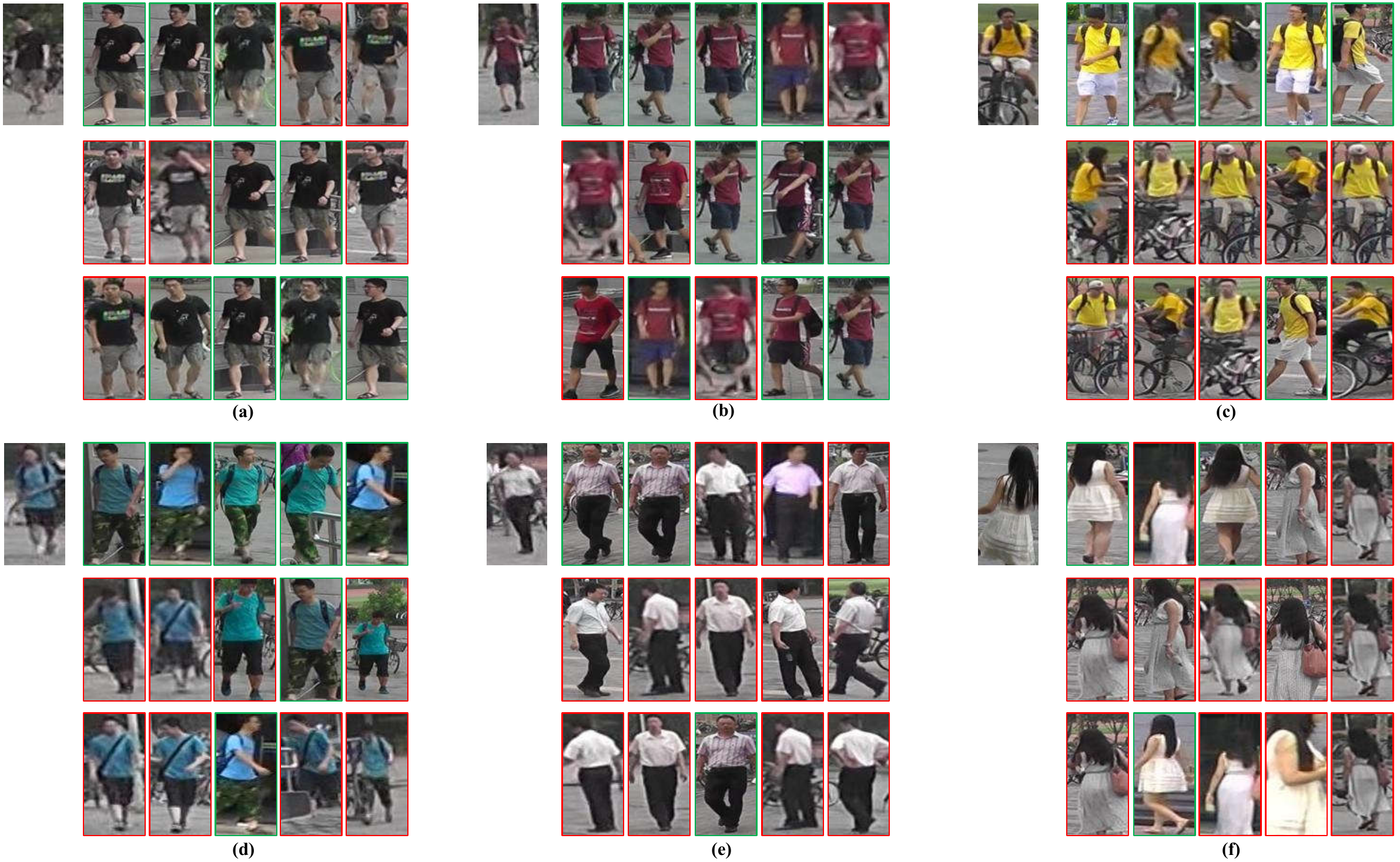}}
  \caption{Examples of the retrieval results on the Market-1501 database. In each group of images, the leftmost one represents the query image, while the remaining images in the first, second, and third rows are the top-5 retrieval results from CDPM, PCB \cite{sun2017beyond}, and MGN \cite{wang2018learning}, respectively. Green rectangles denote true positives, while red rectangles indicate the false positives. Best viewed in color.}
  \label{img13}
  }
\end{figure*}

As illustrated in Fig. \ref{img13}, CDPM is able to yield more reliable results than both PCB and MGN for images with part misalignment problems (Fig. \ref{img13}(a)(b)(d)(e)); this is due to its ability to flexibly align the body parts between images. Moreover, we can also observe that CDPM is more robust to both occlusion (Fig. \ref{img13}(c)) and part-missing problems (Fig. \ref{img13}(f)) than the other two approaches. These results further demonstrate the effectiveness of the proposed CDPM.

\section{Conclusion}
In this work, we study the part misalignment problem in ReID and propose a novel framework named CDPM, which integrates part-level feature learning and part alignment in a single succinct model. In a departure from existing works, we decouple the complicated part misalignment problem into two orthogonal and sequential steps: the first step detects body parts in the vertical direction, while the second step separately refines the boundary of each body part in the horizontal direction. Thanks to the \emph{divide-and-conquer} strategy, each of these two steps becomes significantly simpler. We conduct extensive experiments on three large-scale ReID benchmarks, through which the effectiveness of the proposed model is comprehensively justified and state-of-the-art performance is achieved. We also conduct detailed ablation study to prove the effectiveness of each component in the proposed model.

\bibliographystyle{IEEEtran}
\bibliography{egbib}

\begin{thebibliography}{10}
\providecommand{\url}[1]{#1}
\csname url@samestyle\endcsname
\providecommand{\newblock}{\relax}
\providecommand{\bibinfo}[2]{#2}
\providecommand{\BIBentrySTDinterwordspacing}{\spaceskip=0pt\relax}
\providecommand{\BIBentryALTinterwordstretchfactor}{4}
\providecommand{\BIBentryALTinterwordspacing}{\spaceskip=\fontdimen2\font plus
\BIBentryALTinterwordstretchfactor\fontdimen3\font minus
  \fontdimen4\font\relax}
\providecommand{\BIBforeignlanguage}[2]{{%
\expandafter\ifx\csname l@#1\endcsname\relax
\typeout{** WARNING: IEEEtran.bst: No hyphenation pattern has been}%
\typeout{** loaded for the language `#1'. Using the pattern for}%
\typeout{** the default language instead.}%
\else
\language=\csname l@#1\endcsname
\fi
#2}}
\providecommand{\BIBdecl}{\relax}
\BIBdecl

\bibitem{ristani2018features}
E.~Ristani and C.~Tomasi, ``Features for multi-target multi-camera tracking and
  re-identification,'' in \emph{Proc. IEEE Conf. Comput. Vis. Pattern
  Recognit.}, 2018, pp. 6036--6046.

\bibitem{zheng2017person}
L.~Zheng, H.~Zhang, S.~Sun, M.~Chandraker, Y.~Yang, and Q.~Tian, ``Person
  re-identification in the wild,'' in \emph{Proc. IEEE Conf. Comput. Vis.
  Pattern Recognit.}, 2017, pp. 1367--1376.

\bibitem{zheng2016person}
L.~Zheng, Y.~Yang, and A.~G. Hauptmann, ``Person re-identification: Past,
  present and future,'' \emph{arXiv preprint arXiv:1610.02984}, 2016.

\bibitem{Yao2017Deep}
H.~Yao, S.~Zhang, Y.~Zhang, J.~Li, and Q.~Tian, ``Deep representation learning
  with part loss for person re-identification,'' \emph{IEEE Trans. Image
  Process.}, 2019.

\bibitem{wang2018learning}
G.~Wang, Y.~Yuan, X.~Chen, J.~Li, and X.~Zhou, ``Learning discriminative
  features with multiple granularities for person re-identification,'' in
  \emph{Proc. ACM Int. Conf. Multimedia}, 2018, pp. 274--282.

\bibitem{li2018harmonious}
W.~Li, X.~Zhu, and S.~Gong, ``Harmonious attention network for person
  re-identification,'' in \emph{Proc. IEEE Conf. Comput. Vis. Pattern
  Recognit.}, 2018, pp. 2285--2294.

\bibitem{sun2017beyond}
Y.~Sun, L.~Zheng, Y.~Yang, Q.~Tian, and S.~Wang, ``Beyond part models: Person
  retrieval with refined part pooling (and a strong convolutional baseline),''
  in \emph{Proc. Eur. Conf. Comput. Vis.}, 2018, pp. 480--496.

\bibitem{fu2018horizontal}
Y.~Fu, Y.~Wei, Y.~Zhou, H.~Shi, G.~Huang, X.~Wang, Z.~Yao, and T.~Huang,
  ``Horizontal pyramid matching for person re-identification,'' in \emph{Proc.
  AAAI}, 2019.

\bibitem{Part-based2017tip}
F.~Zhu, X.~Kong, L.~Zheng, H.~Fu, and Q.~Tian, ``Part-based deep hashing for
  large-scale person re-identification,'' \emph{IEEE Trans. Image Process.},
  vol.~26, no.~10, pp. 4806--4817, 2017.

\bibitem{zhang2017alignedreid}
X.~Zhang, H.~Luo, X.~Fan, W.~Xiang, Y.~Sun, Q.~Xiao, W.~Jiang, C.~Zhang, and
  J.~Sun, ``Alignedreid: Surpassing human-level performance in person
  re-identification,'' \emph{arXiv preprint arXiv:1711.08184}, 2017.

\bibitem{zheng2018pedestrian}
Z.~Zheng, L.~Zheng, and Y.~Yang, ``Pedestrian alignment network for large-scale
  person re-identification,'' \emph{IEEE Trans. Syst., Man, Cybern., Syst.},
  2018.

\bibitem{zhao2017spindle}
H.~Zhao, M.~Tian, S.~Sun, J.~Shao, J.~Yan, S.~Yi, X.~Wang, and X.~Tang,
  ``Spindle net: Person re-identification with human body region guided feature
  decomposition and fusion,'' in \emph{Proc. IEEE Conf. Comput. Vis. Pattern
  Recognit.}, 2017, pp. 1077--1085.

\bibitem{su2017pose}
C.~Su, J.~Li, S.~Zhang, J.~Xing, W.~Gao, and Q.~Tian, ``Pose-driven deep
  convolutional model for person re-identification,'' in \emph{Proc. IEEE Int.
  Conf. Comput. Vis.}, 2017, pp. 3960--3969.

\bibitem{li2017learning}
D.~Li, X.~Chen, Z.~Zhang, and K.~Huang, ``Learning deep context-aware features
  over body and latent parts for person re-identification,'' in \emph{Proc.
  IEEE Conf. Comput. Vis. Pattern Recognit.}, 2017, pp. 384--393.

\bibitem{lan2017deep}
X.~Lan, H.~Wang, S.~Gong, and X.~Zhu, ``Deep reinforcement learning attention
  selection for person re-identification,'' in \emph{Proc. Bri. Mach. Vis.
  Conf.}, 2017, pp. 4--7.

\bibitem{zhao2017deeply}
L.~Zhao, X.~Li, Y.~Zhuang, and J.~Wang, ``Deeply-learned part-aligned
  representations for person re-identification,'' in \emph{Proc. IEEE Int.
  Conf. Comput. Vis.}, 2017, pp. 3219--3228.

\bibitem{liu2017hydraplus}
X.~Liu, H.~Zhao, M.~Tian, L.~Sheng, J.~Shao, S.~Yi, J.~Yan, and X.~Wang,
  ``Hydraplus-net: Attentive deep features for pedestrian analysis,'' in
  \emph{Proc. IEEE Int. Conf. Comput. Vis.}, 2017, pp. 350--359.

\bibitem{zhang2018occluded}
S.~Zhang, J.~Yang, and B.~Schiele, ``Occluded pedestrian detection through
  guided attention in cnns,'' in \emph{Proc. IEEE Conf. Comput. Vis. Pattern
  Recognit.}, 2018, pp. 6995--7003.

\bibitem{ding2017trunk}
C.~Ding and D.~Tao, ``Trunk-branch ensemble convolutional neural networks for
  video-based face recognition,'' \emph{IEEE Trans. Pattern Anal. Mach.
  Intell.}, vol.~40, no.~4, pp. 1002--1014, 2018.

\bibitem{Zheng2015Scalable}
L.~Zheng, L.~Shen, L.~Tian, S.~Wang, J.~Wang, and Q.~Tian, ``Scalable person
  re-identification: A benchmark,'' in \emph{Proc. IEEE Int. Conf. Comput.
  Vis.}, 2015, pp. 1116--1124.

\bibitem{zheng2017unlabeled}
Z.~Zheng, L.~Zheng, and Y.~Yang, ``Unlabeled samples generated by gan improve
  the person re-identification baseline in vitro,'' in \emph{Proc. IEEE Int.
  Conf. Comput. Vis.}, 2017, pp. 3754--3762.

\bibitem{li2014deepreid}
W.~Li, R.~Zhao, T.~Xiao, and X.~Wang, ``Deepreid: Deep filter pairing neural
  network for person re-identification,'' in \emph{Proc. IEEE Conf. Comput.
  Vis. Pattern Recognit.}, 2014, pp. 152--159.

\bibitem{varior2016learning}
R.~R. Varior, G.~Wang, J.~Lu, and T.~Liu, ``Learning invariant color features
  for person reidentification,'' \emph{IEEE Trans. Image Process.}, vol.~25,
  no.~7, pp. 3395--3410, 2016.

\bibitem{zhao2013unsupervised}
R.~Zhao, W.~Ouyang, and X.~Wang, ``Unsupervised salience learning for person
  re-identification,'' in \emph{Proc. IEEE Conf. Comput. Vis. Pattern
  Recognit.}, 2013, pp. 3586--3593.

\bibitem{chen2015relevance}
J.~Chen, Z.~Zhang, and Y.~Wang, ``Relevance metric learning for person
  re-identification by exploiting listwise similarities,'' \emph{IEEE Trans. on
  image Process.}, vol.~24, no.~12, pp. 4741--4755, 2015.

\bibitem{Discriminant2017tip}
J.~Garcia, N.~Martinel, A.~Gardel, I.~Bravo, G.~L. Foresti, and C.~Micheloni,
  ``Discriminant context information analysis for post-ranking person
  re-identification,'' \emph{IEEE Trans. Image Process.}, vol.~26, no.~4, pp.
  1650--1665, 2017.

\bibitem{Relevance2015tip}
J.~Chen, Z.~Zhang, and Y.~Wang, ``Relevance metric learning for person
  re-identification by exploiting listwise similarities,'' \emph{IEEE Trans.
  Image Process.}, vol.~24, no.~12, pp. 4741--4755, 2015.

\bibitem{kernel2019tip}
B.~Nguyen and B.~De~Baets, ``Kernel distance metric learning using pairwise
  constraints for person re-identification,'' \emph{IEEE Trans. Image
  Process.}, vol.~28, no.~2, pp. 589--600, 2019.

\bibitem{metric2018tip}
X.~Yang, M.~Wang, and D.~Tao, ``Person re-identification with metric learning
  using privileged information,'' \emph{IEEE Trans. Image Process.}, vol.~27,
  no.~2, pp. 791--805, 2018.

\bibitem{newxiao2016learning}
T.~Xiao, H.~Li, W.~Ouyang, and X.~Wang, ``Learning deep feature representations
  with domain guided dropout for person re-identification,'' in \emph{Proc.
  IEEE Conf. Comput. Vis. Pattern Recognit.}, 2016, pp. 1249--1258.

\bibitem{newchen2018group}
D.~Chen, D.~Xu, H.~Li, N.~Sebe, and X.~Wang, ``Group consistent similarity
  learning via deep crf for person re-identification,'' in \emph{Proc. IEEE
  Conf. Comput. Vis. Pattern Recognit.}, 2018, pp. 8649--8658.

\bibitem{shen2018end}
Y.~Shen, T.~Xiao, H.~Li, S.~Yi, and X.~Wang, ``End-to-end deep
  kronecker-product matching for person re-identification,'' in \emph{Proc.
  IEEE Conf. Comput. Vis. Pattern Recognit.}, 2018, pp. 6886--6895.

\bibitem{Learning2018tip}
Z.~Feng, J.~Lai, and X.~Xie, ``Learning view-specific deep networks for person
  re-identification,'' \emph{IEEE Trans. Image Process.}, vol.~27, no.~7, pp.
  3472--3483, 2018.

\bibitem{Video2019tip}
J.~Dai, P.~Zhang, D.~Wang, H.~Lu, and H.~Wang, ``Video person re-identification
  by temporal residual learning,'' \emph{IEEE Trans. Image Process.}, vol.~28,
  no.~3, pp. 1366--1377, 2019.

\bibitem{Robust2017tip}
A.~Wu, W.-S. Zheng, and J.-H. Lai, ``Robust depth-based person
  re-identification,'' \emph{IEEE Trans. Image Process.}, vol.~26, no.~6, pp.
  2588--2603, 2017.

\bibitem{Ranking2016tip}
S.-Z. Chen, C.-C. Guo, and J.-H. Lai, ``Deep ranking for person
  re-identification via joint representation learning,'' \emph{IEEE Trans.
  Image Process.}, vol.~25, no.~5, pp. 2353--2367, 2016.

\bibitem{shen2018deep}
Y.~Shen, H.~Li, T.~Xiao, S.~Yi, D.~Chen, and X.~Wang, ``Deep group-shuffling
  random walk for person re-identification,'' in \emph{Proc. IEEE Conf. Comput.
  Vis. Pattern Recognit.}, 2018, pp. 2265--2274.

\bibitem{End2017tip}
H.~Liu, J.~Feng, M.~Qi, J.~Jiang, and S.~Yan, ``End-to-end comparative
  attention networks for person re-identification,'' \emph{IEEE Trans. Image
  Process.}, vol.~26, no.~7, pp. 3492--3506, 2017.

\bibitem{newwang2018person}
Y.~Wang, Z.~Chen, F.~Wu, and G.~Wang, ``Person re-identification with cascaded
  pairwise convolutions,'' in \emph{Proc. IEEE Conf. Comput. Vis. Pattern
  Recognit.}, 2018, pp. 1470--1478.

\bibitem{deng2018image}
W.~Deng, L.~Zheng, Q.~Ye, G.~Kang, Y.~Yang, and J.~Jiao, ``Image-image domain
  adaptation with preserved self-similarity and domain-dissimilarity for person
  re-identification,'' in \emph{Proc. IEEE Conf. Comput. Vis. Pattern
  Recognit.}, 2018, pp. 994--1003.

\bibitem{Varior2016A}
R.~R. Varior, B.~Shuai, J.~Lu, D.~Xu, and G.~Wang, ``A siamese long short-term
  memory architecture for human re-identification,'' in \emph{Proc. Eur. Conf.
  Comput. Vis.}, 2016, pp. 135--153.

\bibitem{hermans2017defense}
A.~Hermans, L.~Beyer, and B.~Leibe, ``In defense of the triplet loss for person
  re-identification,'' \emph{arXiv preprint arXiv:1703.07737}, 2017.

\bibitem{chen2017beyond}
W.~Chen, X.~Chen, J.~Zhang, and K.~Huang, ``Beyond triplet loss: a deep
  quadruplet network for person re-identification,'' in \emph{Proc. IEEE Conf.
  Comput. Vis. Pattern Recognit.}, 2017, pp. 403--412.

\bibitem{macus2018}
C.~Wang, Q.~Zhang, C.~Huang, W.~Liu, and X.~Wang, ``Mancs: A multi-task
  attentional network with curriculum sampling for person re-identification,''
  in \emph{Proc. Eur. Conf. Comput. Vis.}, 2018, pp. 365--381.

\bibitem{chen2017multi}
W.~Chen, X.~Chen, J.~Zhang, and K.~Huang, ``A multi-task deep network for
  person re-identification,'' in \emph{Proc. AAAI}, 2017.

\bibitem{ahmed2015improved}
E.~Ahmed, M.~Jones, and T.~K. Marks, ``An improved deep learning architecture
  for person re-identification,'' in \emph{Proc. IEEE Conf. Comput. Vis.
  Pattern Recognit.}, 2015, pp. 3908--3916.

\bibitem{Cheng2016Person}
D.~Cheng, Y.~Gong, S.~Zhou, J.~Wang, and N.~Zheng, ``Person re-identification
  by multi-channel parts-based cnn with improved triplet loss function,'' in
  \emph{Proc. IEEE Conf. Comput. Vis. Pattern Recognit.}, pp. 1335--1344.

\bibitem{song2018mask}
C.~Song, Y.~Huang, W.~Ouyang, and L.~Wang, ``Mask-guided contrastive attention
  model for person re-identification,'' in \emph{Proc. IEEE Conf. Comput. Vis.
  Pattern Recognit.}, 2018, pp. 1179--1188.

\bibitem{tian2018eliminating}
M.~Tian, S.~Yi, H.~Li, S.~Li, X.~Zhang, J.~Shi, J.~Yan, and X.~Wang,
  ``Eliminating background-bias for robust person re-identification,'' in
  \emph{Proc. IEEE Conf. Comput. Vis. Pattern Recognit.}, 2018, pp. 5794--5803.

\bibitem{kalayeh2018human}
M.~M. Kalayeh, E.~Basaran, M.~G{\"o}kmen, M.~E. Kamasak, and M.~Shah, ``Human
  semantic parsing for person re-identification,'' in \emph{Proc. IEEE Conf.
  Comput. Vis. Pattern Recognit.}, 2018, pp. 1062--1071.

\bibitem{xu2018attention}
J.~Xu, R.~Zhao, F.~Zhu, H.~Wang, and W.~Ouyang, ``Attention-aware compositional
  network for person re-identification,'' in \emph{Proc. IEEE Conf. Comput.
  Vis. Pattern Recognit.}, 2018, pp. 2119--2128.

\bibitem{felzenszwalb2008discriminatively}
P.~Felzenszwalb, D.~McAllester, and D.~Ramanan, ``A discriminatively trained,
  multiscale, deformable part model,'' in \emph{Proc. IEEE Conf. Comput. Vis.
  Pattern Recognit.}, 2008.

\bibitem{ouyang2015deepid}
W.~Ouyang, X.~Wang, X.~Zeng, S.~Qiu, P.~Luo, Y.~Tian, H.~Li, S.~Yang, Z.~Wang,
  C.-C. Loy \emph{et~al.}, ``Deepid-net: Deformable deep convolutional neural
  networks for object detection,'' in \emph{Proc. IEEE Conf. Comput. Vis.
  Pattern Recognit.}, 2015, pp. 2403--2412.

\bibitem{savalle2014deformable}
P.-A. Savalle, S.~Tsogkas, G.~Papandreou, and I.~Kokkinos, ``Deformable part
  models with cnn features,'' in \emph{Proc. Eur. Conf. Comput. Vis.
  Workshops}, 2014.

\bibitem{girshick2015deformable}
R.~Girshick, F.~Iandola, T.~Darrell, and J.~Malik, ``Deformable part models are
  convolutional neural networks,'' in \emph{Proc. IEEE Conf. Comput. Vis.
  Pattern Recognit.}, 2015, pp. 437--446.

\bibitem{girshick2015fast}
R.~Girshick, ``Fast r-cnn,'' in \emph{Proc. IEEE Int. Conf. Comput. Vis.},
  2015, pp. 1440--1448.

\bibitem{ren2017faster}
S.~Ren, K.~He, R.~Girshick, and J.~Sun, ``Faster r-cnn: Towards real-time
  object detection with region proposal networks,'' \emph{IEEE Trans. Pattern
  Anal. Mach. Intell.}, vol.~39, no.~6, p. 1137, 2017.

\bibitem{luo2018macro}
Y.~Luo, Z.~Zheng, L.~Zheng, T.~Guan, J.~Yu, and Y.~Yang, ``Macro-micro
  adversarial network for human parsing,'' in \emph{Proc. IEEE Conf. Comput.
  Vis. Pattern Recognit.}, 2018, pp. 418--434.

\bibitem{vapnik2015learning}
V.~Vapnik and R.~Izmailov, ``Learning using privileged information: similarity
  control and knowledge transfer,'' \emph{J. Mach. Learn. Res.}, vol.~16, no.
  2023-2049, p.~2, 2015.

\bibitem{he2016deep}
K.~He, X.~Zhang, S.~Ren, and J.~Sun, ``Deep residual learning for image
  recognition,'' in \emph{Proc. IEEE Conf. Comput. Vis. Pattern Recognit.},
  2016, pp. 770--778.

\bibitem{hu2018senet}
J.~Hu, L.~Shen, and G.~Sun, ``Squeeze-and-excitation networks,'' in \emph{Proc.
  IEEE Conf. Comput. Vis. Pattern Recognit.}, 2018, pp. 7132--7141.

\bibitem{zhong2017re}
Z.~Zhong, L.~Zheng, D.~Cao, and S.~Li, ``Re-ranking person re-identification
  with k-reciprocal encoding,'' in \emph{Proc. IEEE Conf. Comput. Vis. Pattern
  Recognit.}, 2017, pp. 1318--1327.

\bibitem{zhong2017random}
Z.~Zhong, L.~Zheng, G.~Kang, S.~Li, and Y.~Yang, ``Random erasing data
  augmentation,'' \emph{arXiv preprint arXiv:1708.04896}, 2017.

\bibitem{sutskever2013importance}
I.~Sutskever, J.~Martens, G.~E. Dahl, and G.~E. Hinton, ``On the importance of
  initialization and momentum in deep learning,'' \emph{Proc. Int. Conf. Mach.
  Learn.}, vol.~28, no. 1139-1147, p.~5.

\bibitem{sun2017svdnet}
Y.~Sun, L.~Zheng, W.~Deng, and S.~Wang, ``Svdnet for pedestrian retrieval,'' in
  \emph{Proc. IEEE Int. Conf. Comput. Vis.}, 2017, pp. 3800--3808.

\bibitem{wang2018resource}
Y.~Wang, L.~Wang, Y.~You, X.~Zou, V.~Chen, S.~Li, G.~Huang, B.~Hariharan, and
  K.~Q. Weinberger, ``Resource aware person re-identification across multiple
  resolutions,'' in \emph{Proc. IEEE Conf. Comput. Vis. Pattern Recognit.},
  2018, pp. 8042--8051.

\bibitem{chang2018multi}
X.~Chang, T.~M. Hospedales, and T.~Xiang, ``Multi-level factorisation net for
  person re-identification,'' in \emph{Proc. IEEE Conf. Comput. Vis. Pattern
  Recognit.}, 2018, pp. 2109--2118.

\bibitem{suh2018part}
Y.~Suh, J.~Wang, S.~Tang, T.~Mei, and K.~M. Lee, ``Part-aligned bilinear
  representations for person re-identification,'' in \emph{Proc. Eur. Conf.
  Comput. Vis.}, 2018, pp. 402--419.

\bibitem{chen2017person}
Y.~Chen, X.~Zhu, and S.~Gong, ``Person re-identification by deep learning
  multi-scale representations,'' in \emph{Proc. IEEE Int. Conf. Comput. Vis.},
  2017, pp. 2590--2600.

\bibitem{zhang2018person62}
Z.~Zhang and M.~Huang, ``Person re-identification based on heterogeneous
  part-based deep network in camera networks,'' \emph{IEEE Trans. Emerg. Topics
  in Comput. Intell.}, 2018.

\bibitem{maaten2008visualizing}
L.~v.~d. Maaten and G.~Hinton, ``Visualizing data using t-sne,'' \emph{Journal
  of machine learning research}, pp. 2579--2605, 2008.

\bibitem{selvaraju2017grad}
R.~R. Selvaraju, M.~Cogswell, A.~Das, R.~Vedantam, D.~Parikh, and D.~Batra,
  ``Grad-cam: Visual explanations from deep networks via gradient-based
  localization,'' in \emph{Proc. IEEE Int. Conf. Comput. Vis.}, 2017, pp.
  618--626.

\end{thebibliography}

\end{document}